\documentclass[10pt,journal,cspaper,compsoc]{IEEEtran}

\usepackage{ifpdf}
\pdfoutput=1
\usepackage[nocompress]{cite}
\usepackage{graphicx}
\usepackage{algorithmic}
\usepackage{color}
\usepackage{subfigure}
\usepackage{placeins}
\usepackage{url}

\newenvironment{packed_itemize}{
\vspace{-0.15cm}\begin{itemize}
  \setlength{\itemsep}{1pt}
  \setlength{\parskip}{0pt}
  \setlength{\parsep}{0pt}
}{\end{itemize}}

\newenvironment{packed_enumerate}{
\begin{enumerate}
  \setlength{\itemsep}{1pt}
  \setlength{\parskip}{0pt}
  \setlength{\parsep}{0pt}
}{\end{enumerate}}

\newcommand{\etal}{\emph{et al.}}
\newcommand{\eg}{\emph{e.g.,}}
\newcommand{\ie}{\emph{i.e.,}}

\begin{document}
%

\title{Image Super-Resolution Using Deep Convolutional Networks}

\author{Chao~Dong,
        Chen~Change~Loy,~\IEEEmembership{Member,~IEEE,}
        Kaiming~He,~\IEEEmembership{Member,~IEEE,}
        and~Xiaoou~Tang,~\IEEEmembership{Fellow,~IEEE}
\IEEEcompsocitemizethanks{\IEEEcompsocthanksitem C. Dong, C. C. Loy and X. Tang are with the Department of Information Engineering, The Chinese University of Hong Kong, Hong Kong.\protect\\
E-mail: \{dc012,ccloy,xtang\}@ie.cuhk.edu.hk
\IEEEcompsocthanksitem K. He is with the Visual Computing Group, Microsoft Research Asia, Beijing 100080, China. \protect\\
Email: kahe@microsoft.com}
\thanks{}}

\markboth{}%
{Shell \MakeLowercase{\textit{et al.}}: Bare Demo of IEEEtran.cls for Computer Society Journals}

\IEEEcompsoctitleabstractindextext{%
\begin{abstract}

We propose a deep learning method for single image super-resolution (SR). Our method directly learns an end-to-end mapping between the low/high-resolution images.
The mapping is represented as a deep convolutional neural network (CNN) that takes the low-resolution image as the input and outputs the high-resolution one.
We further show that traditional sparse-coding-based SR methods can also be viewed as a deep convolutional network. But unlike traditional methods that handle each component separately, our method jointly optimizes all layers.
Our deep CNN has a lightweight structure, yet demonstrates state-of-the-art restoration quality, and achieves fast speed for practical on-line usage.
We explore different network structures and parameter settings to achieve trade-offs between performance and speed.
Moreover, we extend our network to cope with three color channels simultaneously, and show better overall reconstruction quality.

\end{abstract}
\begin{keywords}
Super-resolution, deep convolutional neural networks, sparse coding
\end{keywords}}

\maketitle

\IEEEdisplaynotcompsoctitleabstractindextext
\IEEEpeerreviewmaketitle

\section{Introduction}

\setcounter{footnote}{0}

Single image super-resolution (SR)~\cite{Irani1991}, which aims at recovering a high-resolution image from a single low-resolution image, is a classical problem in computer vision.
This problem is inherently ill-posed since a multiplicity of solutions exist for any given low-resolution pixel. In other words, it is an underdetermined inverse problem, of which solution is not unique. Such a problem is typically mitigated by constraining the solution space by strong prior information.
To learn the prior, recent state-of-the-art methods mostly adopt the example-based~\cite{Yang2014} strategy.
These methods either exploit internal similarities of the same image~\cite{Freedman2011,Glasner2009,Yang2013,Cui2014,Huang2015}, or learn mapping functions from external low- and high-resolution exemplar pairs \cite{Bevilacqua2012,Chang2004,Freeman2000,jia2013image,Timofte2013,Yang2012,Yang2010a,Zeyde2012,Kim2010,Timofte2014,Yang2013,Dai2015,Schulter2015}.
The external example-based methods can be formulated for generic image super-resolution, or can be designed to suit domain specific tasks, \ie~face hallucination~\cite{liu2007face, Yang2010a}, according to the training samples provided.

The sparse-coding-based method~\cite{Yang2008,Yang2010a} is one of the representative external example-based SR methods. This method involves several steps in its solution pipeline.
First, overlapping patches are densely cropped from the input image and pre-processed (\eg subtracting mean and normalization).
These patches are then encoded by a low-resolution dictionary. The sparse coefficients are passed into a high-resolution dictionary for reconstructing high-resolution patches. The overlapping reconstructed patches are aggregated (\eg~by weighted averaging) to produce the final output. This pipeline is shared by most external example-based methods, which pay particular attention to learning and optimizing the dictionaries \cite{Yang2008,Yang2010a,Bevilacqua2012} or building efficient mapping functions~\cite{Kim2010,Timofte2013,Timofte2014,Yang2013}. However, the rest of the steps in the pipeline have been rarely optimized or considered in an unified optimization framework.

\begin{figure}[t]
\begin{center}

 \includegraphics[width=0.85\linewidth]{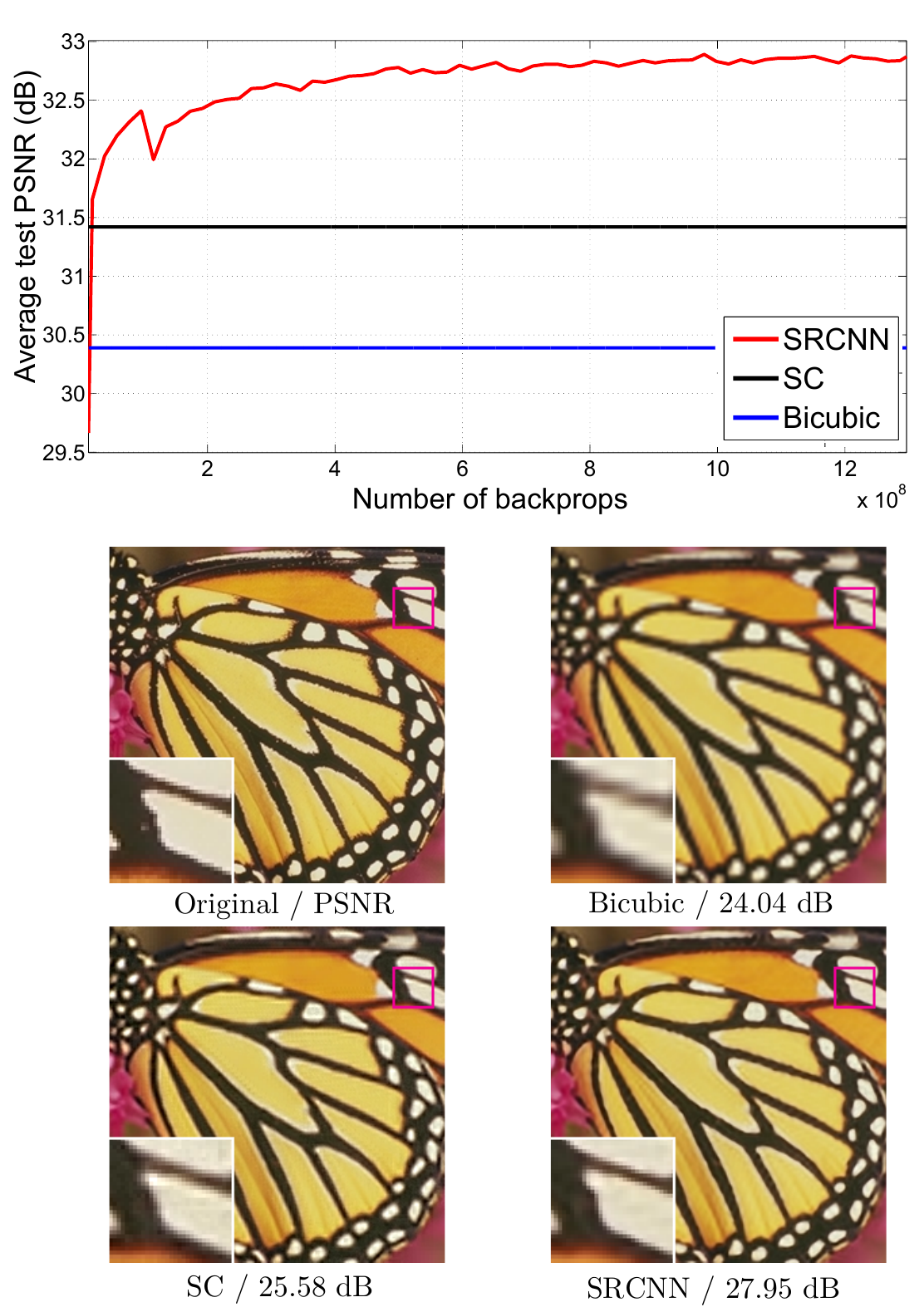}%

\caption{The proposed Super-Resolution Convolutional Neural Network (SRCNN) surpasses the bicubic baseline with just a few training iterations, and outperforms the sparse-coding-based method (SC)~\cite{Yang2010a} with moderate training. The performance may be further improved with more training iterations. More details are provided in Section~\ref{sec:quality} (the Set5 dataset with an upscaling factor 3). The proposed method provides visually appealing reconstructed image.}
\label{fig:overview}
\vspace{-0.25cm}
\end{center}
\vskip -0.4cm
\end{figure}

In this paper, we show that the aforementioned pipeline is equivalent to a deep convolutional neural network \cite{LeCun1989} (more details in Section~\ref{subsec:illustraion}). Motivated by this fact, we consider a convolutional neural network that directly learns an end-to-end mapping between low- and high-resolution images.
Our method differs fundamentally from existing external example-based approaches, in that  ours does not explicitly learn the dictionaries~\cite{Timofte2013,Yang2008,Yang2010a} or manifolds~\cite{Bevilacqua2012,Chang2004} for modeling the patch space. These are implicitly achieved via hidden layers. Furthermore, the patch extraction and aggregation are also formulated as convolutional layers, so are involved in the optimization. In our method, the entire SR pipeline is fully obtained through learning, with little pre/post-processing.

We name the proposed model Super-Resolution Convolutional Neural Network (SRCNN)\footnote{The implementation is available at \url{http://mmlab.ie.cuhk.edu.hk/projects/SRCNN.html}.}. The proposed SRCNN has several appealing properties.
First, its structure is intentionally designed with simplicity in mind, and yet provides superior accuracy\footnote{Numerical evaluations by using different metrics such as the Peak Signal-to-Noise Ratio (PSNR), structure similarity index (SSIM)~\cite{wang2004image}, multi-scale SSIM~\cite{wang2003multiscale}, information fidelity criterion~\cite{sheikh2005information}, when the ground truth images are available.} compared with state-of-the-art example-based methods.
Figure~\ref{fig:overview} shows a comparison on an example.
Second, with moderate numbers of filters and layers, our method achieves fast speed for practical on-line usage even on a CPU. Our method is faster than a number of example-based methods, because it is fully feed-forward and does not need to solve any optimization problem on usage. Third, experiments show that the restoration quality of the network can be further improved when (i) larger and more diverse datasets are available, and/or (ii) a larger and deeper model is used. On the contrary, larger datasets/models can present challenges for existing example-based methods.
Furthermore,
the proposed network can cope with three channels of color images simultaneously to achieve improved super-resolution performance.

Overall, the contributions of this study are mainly in three aspects:
\begin{enumerate}
\item We present a fully convolutional neural network for image super-resolution. The network directly learns an end-to-end mapping between low- and high-resolution images, with little pre/post-processing beyond the optimization.

\item We establish a relationship between our deep-learning-based SR method and the traditional sparse-coding-based SR methods. This relationship provides a guidance for the design of the network structure.

\item We demonstrate that deep learning is useful in the classical computer vision problem of super-resolution, and can achieve good quality and speed.
\end{enumerate}

A preliminary version of this work was presented earlier~\cite{Dong2014}.
The present work adds to the initial version in significant ways. Firstly, we improve the SRCNN by introducing larger filter size in the non-linear mapping layer, and explore deeper structures by adding non-linear mapping layers.
Secondly, we extend the SRCNN to process three color channels (either in YCbCr or RGB color space) simultaneously. Experimentally, we demonstrate that performance can be improved in comparison to the single-channel network.
Thirdly, considerable new analyses and intuitive explanations are added to the initial results. We also extend the original experiments from \textit{Set5}~\cite{Bevilacqua2012} and \textit{Set14}~\cite{Zeyde2012} test images to \textit{BSD200}~\cite{martin2001database} (200 test images). In addition, we compare with a number of recently published methods and confirm that our model still outperforms existing approaches using different evaluation metrics.

\section{Related Work}
\label{sec:related_work}

\subsection{Image Super-Resolution}

According to the image priors, single-image super resolution algorithms can be categorized into four types -- prediction models, edge based methods, image statistical methods and patch based (or example-based) methods. These methods have been thoroughly investigated and evaluated in Yang~\etal's work~\cite{Yang2014}. Among them, the example-based methods~\cite{Kim2010,Timofte2013,Yang2013,Glasner2009} achieve the state-of-the-art performance.

The internal example-based methods exploit the self-similarity property and generate exemplar patches from the input image. It is first proposed in Glasner's work~\cite{Glasner2009}, and several improved variants~\cite{Freedman2011,Yang2011} are proposed to accelerate the implementation.
The external example-based methods~\cite{Freeman2000,Chang2004,Yang2008,Yang2010a,Yang2012,Bevilacqua2012,Zeyde2012,Timofte2013,Dai2015,Schulter2015} learn a mapping between low/high-resolution patches from external datasets.
These studies vary on how to learn a compact dictionary or manifold space to relate low/high-resolution patches, and on how representation schemes can be conducted in such spaces. In the pioneer work of Freeman~\etal~\cite{Freeman2002},
the dictionaries are directly presented as low/high-resolution patch pairs, and the nearest neighbour (NN) of the input patch is found in the low-resolution space, with its corresponding high-resolution patch used for reconstruction.
Chang~\etal~\cite{Chang2004} introduce a manifold embedding technique as an alternative to the NN strategy.
In Yang~\etal's work~\cite{Yang2008,Yang2010a}, the above NN correspondence advances to a more sophisticated sparse coding formulation. Other mapping functions such as kernel regression~\cite{Kim2010}, simple function~\cite{Yang2013}, random forest~\cite{Schulter2015} and anchored neighborhood regression~\cite{Timofte2013,Timofte2014} are proposed to further improve the mapping accuracy and speed. The sparse-coding-based method and its several improvements~\cite{Yang2012,Timofte2013,Timofte2014} are among the state-of-the-art SR methods nowadays. In these methods, the patches are the focus of the optimization; the patch extraction and aggregation steps are considered as pre/post-processing and handled separately.

The majority of SR algorithms~\cite{Freeman2000,Chang2004,Yang2008,Yang2010a,Yang2012,Bevilacqua2012,Zeyde2012,Timofte2013} focus on gray-scale or single-channel image super-resolution. For color images, the aforementioned methods first transform the problem to a different color space (YCbCr or YUV), and SR is applied only on the luminance channel.
There are also works attempting to super-resolve all channels simultaneously. For example, Kim and Kwon~\cite{Kim2010} and Dai~\etal~\cite{Dai2009} apply their model to each RGB channel and combined them to produce the final results. However, none of them has analyzed the SR performance of different channels, and the necessity of recovering all three channels.

\subsection{Convolutional Neural Networks}

Convolutional neural networks (CNN) date back decades~\cite{LeCun1989} and deep CNNs  have recently shown an explosive popularity partially due to its success in image classification~\cite{Krizhevsky2012,He2014}. They have also been successfully applied to other computer vision fields, such as object detection~\cite{Ouyang2014,zhang2014part,szegedy2014scalable}, face recognition~\cite{sun2014deepb}, and pedestrian detection~\cite{ouyang2013joint}. Several factors are of central importance in this progress: (i) the efficient training implementation on modern powerful GPUs \cite{Krizhevsky2012}, (ii) the proposal of the Rectified Linear Unit (ReLU) \cite{Nair2010} which makes convergence much faster while still presents good quality~\cite{Krizhevsky2012}, and (iii) the easy access to an abundance of data (like ImageNet \cite{Deng2009}) for training larger models. Our method also benefits from these progresses.

\subsection{Deep Learning for Image Restoration}
\label{sec:related_work_deep}
There have been a few studies of using deep learning techniques for image restoration.
The multi-layer perceptron (MLP), whose all layers are fully-connected (in contrast to convolutional), is applied for natural image denoising \cite{Burger2012} and post-deblurring denoising \cite{Schuler2013}.
More closely related to our work, the convolutional neural network is applied for natural image denoising \cite{jain2008natural} and removing noisy patterns (dirt/rain) \cite{Eigen2013}.
These restoration problems are more or less denoising-driven.
Cui~\etal~\cite{Cui2014} propose to embed auto-encoder networks in their super-resolution pipeline under the notion internal example-based approach~\cite{Glasner2009}.
The deep model is not specifically designed to be an end-to-end solution, since each layer of the cascade requires independent optimization of the self-similarity search process and the auto-encoder. On the contrary, the proposed SRCNN optimizes an end-to-end mapping. Further, the SRCNN is faster at speed. It is not only a quantitatively superior method, but also a practically useful one.

\section{Convolutional Neural Networks for Super-Resolution}
\label{sec:methodology}

\subsection{Formulation}
\label{subsec:Formulation}

\begin{figure*}[t]
\centering

  \includegraphics[width=0.8\linewidth]{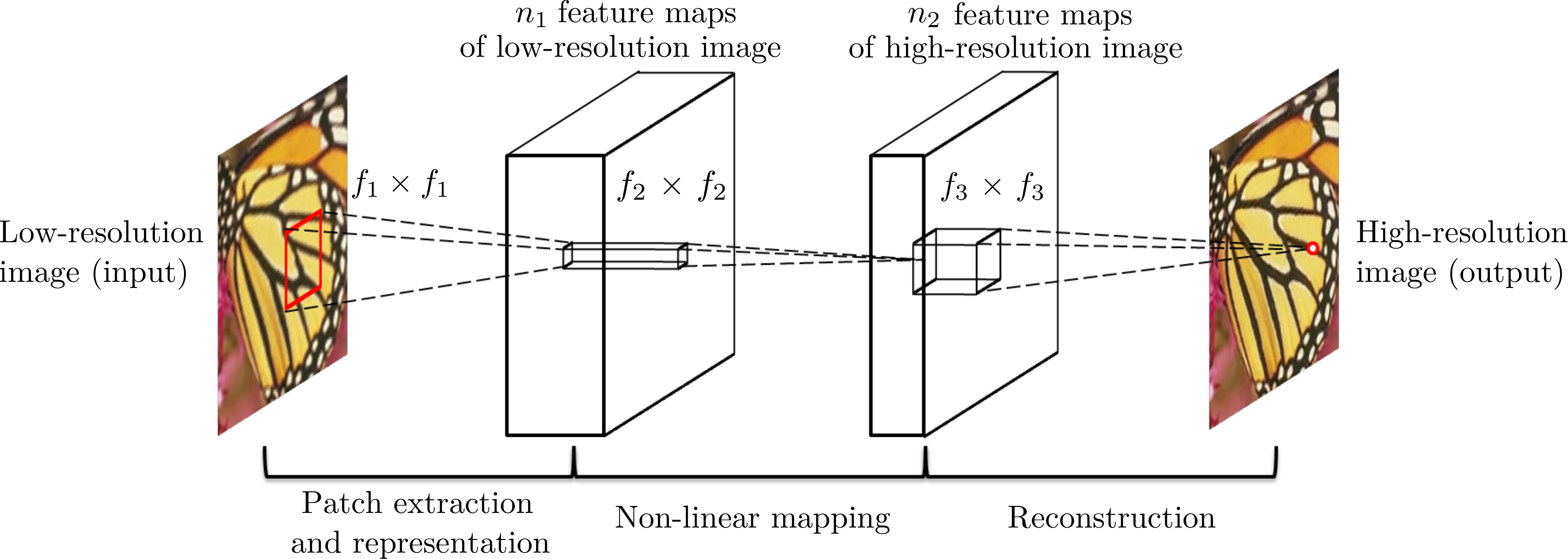}

\vskip -0.25cm
  \caption{Given a low-resolution image $\mathbf{Y}$, the first convolutional layer of the SRCNN extracts a set of feature maps. The second layer maps these feature maps nonlinearly to high-resolution patch representations. The last layer combines the predictions within a spatial neighbourhood to produce the final high-resolution image $F(\mathbf{Y})$.}
  \label{fig:structure}
\vskip -0.4cm
\end{figure*}

Consider a single low-resolution image, we first upscale it to the desired size using bicubic interpolation, which is the only pre-processing we perform\footnote{Bicubic interpolation is also a convolutional operation, so it can be formulated as a convolutional layer. However, the output size of this layer is larger than the input size, so there is a fractional stride. To take advantage of the popular well-optimized implementations such as \emph{cuda-convnet} \cite{Krizhevsky2012}, we exclude this ``layer'' from learning.}.
Let us denote the interpolated image as $\mathbf{Y}$.
Our goal is to recover from $\mathbf{Y}$ an image $F(\mathbf{Y})$ that is as similar as possible to the ground truth high-resolution image $\mathbf{X}$.
For the ease of presentation, we still call $\mathbf{Y}$ a ``low-resolution'' image, although it has the same size as $\mathbf{X}$.
We wish to learn a mapping $F$, which conceptually consists of three operations:
\begin{packed_enumerate}
\item \textbf{Patch extraction and representation:} this operation extracts (overlapping) patches from the low-resolution image $\mathbf{Y}$ and represents each patch as a high-dimensional vector. These vectors comprise a set of feature maps, of which the number equals to the dimensionality of the vectors.
\item \textbf{Non-linear mapping:} this operation nonlinearly maps each high-dimensional vector onto another high-dimensional vector. Each mapped vector is conceptually the representation of a high-resolution patch. These vectors comprise another set of feature maps.
\item \textbf{Reconstruction:} this operation aggregates the above high-resolution patch-wise representations to generate the final high-resolution image. This image is expected to be similar to the ground truth $\mathbf{X}$.
\end{packed_enumerate}
We will show that all these operations form a convolutional neural network. An overview of the network is depicted in Figure~\ref{fig:structure}. Next we detail our definition of each operation.

\subsubsection{Patch extraction and representation}
\label{subsubsec:first_layer}

A popular strategy in image restoration (\eg~\cite{Aharon2006}) is to densely extract patches and then represent them by a set of pre-trained bases such as PCA, DCT, Haar, etc. This is equivalent to convolving the image by a set of filters, each of which is a basis. In our formulation, we involve the optimization of these bases into the optimization of the network. Formally, our first layer is expressed as an operation $F_{1}$:
\begin{equation}
\label{eqn:first_layer}
F_{1}(\mathbf{Y})=\max\left(0, W_{1}*\mathbf{Y}+B_{1}\right),
\end{equation}
where $W_{1}$ and $B_{1}$ represent the filters and biases respectively, and '$*$' denotes the convolution operation.
Here, $W_{1}$ corresponds to $n_1$ filters of support $c\times f_1 \times f_1$, where $c$ is the number of channels in the input image, $f_1$ is the spatial size of a filter. Intuitively, $W_{1}$ applies $n_1$ convolutions on the image, and each convolution has a kernel size $c\times f_1 \times f_1$. The output is composed of $n_1$ feature maps. $B_{1}$ is an $n_1$-dimensional vector, whose each element is associated with a filter.
%
We apply the Rectified Linear Unit (ReLU, $\max(0,x)$) \cite{Nair2010} on the filter responses\footnote{The ReLU can be equivalently considered as a part of the second operation (Non-linear mapping), and the first operation (Patch extraction and representation) becomes purely linear convolution.}.

\subsubsection{Non-linear mapping}
\label{subsubsec:second_layer}

The first layer extracts an $n_1$-dimensional feature for each patch.
In the second operation, we map each of these $n_1$-dimensional vectors into an $n_2$-dimensional one. This is equivalent to applying $n_2$ filters which have a trivial spatial support $1 \times 1$.
This interpretation is only valid for $1 \times 1$ filters. But it is easy to generalize to larger filters like $3\times 3$ or $5\times 5$. In that case, the non-linear mapping is not on a patch of the input image; instead, it is on a $3\times 3$ or $5\times 5$ ``patch'' of the feature map. The operation of the second layer is:
\begin{equation}
\label{eqn:second_layer}
F_{2}(\mathbf{Y})=\max\left(0, W_{2}*F_{1}(\mathbf{Y})+B_{2}\right).
\end{equation}
Here $W_{2}$ contains $n_2$ filters of size $n_1\times f_2 \times f_2$, and $B_{2}$ is $n_2$-dimensional.
Each of the output $n_2$-dimensional vectors is conceptually a representation of a high-resolution patch that will be used for reconstruction.

It is possible to add more convolutional layers to increase the non-linearity. But this can increase the complexity of the model ($n_2 \times f_2 \times f_2 \times n_2$ parameters for one layer), and thus demands more training time. We will explore deeper structures by introducing additional non-linear mapping layers in Section~\ref{sec:Deeper}.

\subsubsection{Reconstruction}
\label{subsubsec:third_layer}

In the traditional methods, the predicted overlapping high-resolution patches are often averaged to produce the final full image. The averaging can be considered as a pre-defined filter on a set of feature maps (where each position is the ``flattened'' vector form of a high-resolution patch). Motivated by this, we define a convolutional layer to produce the final high-resolution image:
\begin{equation}
\label{eqn:third_lyaer}
F(\mathbf{Y})=W_3*F_{2}(\mathbf{Y})+B_3.
\end{equation}
Here $W_{3}$ corresponds to $c$ filters of a size $n_2\times f_3 \times f_3$, and $B_3$ is a $c$-dimensional vector.

\begin{figure*}[t]
\centering
  \includegraphics[width=0.8\linewidth]{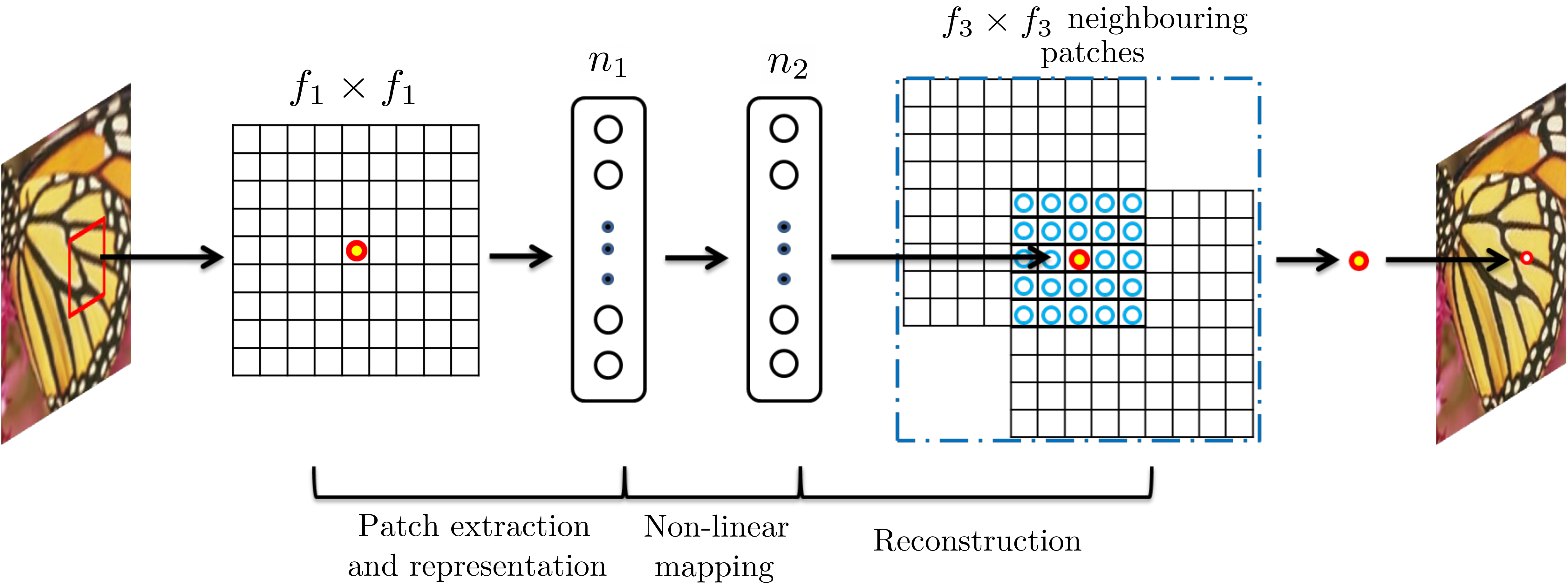}
  \vskip -0.25cm
  \caption{An illustration of sparse-coding-based methods in the view of a convolutional neural network.}
  \label{fig:illustration}
\vskip -0.4cm
\end{figure*}

If the representations of the high-resolution patches are in the image domain (\ie we can simply reshape each representation to form the patch), we expect that the filters act like an averaging filter; if the representations of the high-resolution patches are in some other domains (\eg coefficients in terms of some bases), we expect that $W_{3}$ behaves like first projecting the coefficients onto the image domain and then averaging. In either way, $W_{3}$ is a set of linear filters.

~\\
\indent Interestingly, although the above three operations are motivated by different intuitions, they all lead to the same form as a convolutional layer.
We put all three operations together and form a convolutional neural network (Figure~\ref{fig:structure}).
In this model, all the filtering weights and biases are to be optimized.
Despite the succinctness of the overall structure, our SRCNN model is carefully developed by drawing extensive experience resulted from significant progresses in super-resolution~\cite{Yang2008,Yang2010a}. We detail the relationship in the next section.

\subsection{Relationship to Sparse-Coding-Based Methods}
\label{subsec:illustraion}

We show that the sparse-coding-based SR methods \cite{Yang2008,Yang2010a} can be viewed as a convolutional neural network. Figure~\ref{fig:illustration} shows an illustration.

In the sparse-coding-based methods, let us consider that an $f_1\times f_1$ low-resolution patch is extracted from the input image.
Then the sparse coding solver, like Feature-Sign~\cite{Lee2007}, will first project the patch onto a (low-resolution) dictionary. If the dictionary size is $n_1$, this is equivalent to applying $n_1$ linear filters ($f_1\times f_1$) on the input image (the mean subtraction is also a linear operation so can be absorbed). This is illustrated as the left part of Figure~\ref{fig:illustration}.

The sparse coding solver will then iteratively process the $n_1$ coefficients. The outputs of this solver are $n_2$ coefficients, and usually $n_2=n_1$ in the case of sparse coding. These $n_2$ coefficients are the representation of the high-resolution patch.
In this sense, the sparse coding solver behaves as a special case of a non-linear mapping operator, whose spatial support is $1\times1$. See the middle part of Figure~\ref{fig:illustration}. However, the sparse coding solver is not feed-forward, \ie it is an iterative algorithm.
On the contrary, our non-linear operator is fully feed-forward and can be computed efficiently. If we set $f_2=1$, then our non-linear operator can be considered as a pixel-wise fully-connected layer. It is worth noting that ``the sparse coding solver'' in SRCNN refers to the first two layers, but not just the second layer or the activation function (ReLU). Thus the nonlinear operation in SRCNN is also well optimized through the learning process.

The above $n_2$ coefficients (after sparse coding) are then projected onto another (high-resolution) dictionary to produce a high-resolution patch. The overlapping high-resolution patches are then averaged. As discussed above, this is equivalent to linear convolutions on the $n_2$ feature maps. If the high-resolution patches used for reconstruction are of size $f_3\times f_3$, then the linear filters have an equivalent spatial support of size $f_3\times f_3$. See the right part of Figure~\ref{fig:illustration}.

The above discussion shows that the sparse-coding-based SR method can be viewed as a kind of convolutional neural network (with a different non-linear mapping). But not all operations have been considered in the optimization in the sparse-coding-based SR methods. On the contrary,
in our convolutional neural network, the low-resolution dictionary, high-resolution dictionary, non-linear mapping, together with mean subtraction and averaging, are all involved in the filters to be optimized.
So our method optimizes an end-to-end mapping that consists of all operations.

The above analogy can also help us to design hyper-parameters. For example, we can set the filter size of the last layer to be smaller than that of the first layer, and thus we rely more on the central part of the high-resolution patch (to the extreme, if $f_3=1$, we are using the center pixel with no averaging). We can also set $n_2 < n_1$ because it is expected to be sparser.
A typical and basic setting is $f_1=9$, $f_2=1$, $f_3=5$, $n_1=64$, and $n_2=32$ (we evaluate more settings in the experiment section).
On the whole, the estimation of a high resolution pixel utilizes the information of $(9+5-1)^2=169$ pixels. Clearly, the information exploited for reconstruction is comparatively larger than that used in existing external example-based approaches, \eg~using $(5+5-1)^2=81$ pixels\footnote{The patches are overlapped with 4 pixels at each direction.}~\cite{Freeman2000,Yang2010a}.
This is one of the reasons why the SRCNN gives superior performance.

\subsection{Training}
\label{subsec:learning}

Learning the end-to-end mapping function $F$ requires the estimation of network parameters $\Theta=\{W_1,W_2,W_3,B_1,B_2,B_3\}$. This is achieved through minimizing the loss between the reconstructed images $F(\mathbf{Y} ; \Theta)$ and the corresponding ground truth high-resolution images $\mathbf{X}$.
Given a set of high-resolution images $\left\{ \mathbf{X}_i\right\}$ and their corresponding low-resolution images $\left\{ \mathbf{Y}_i\right\}$, we use Mean Squared Error (MSE) as the loss function:
\begin{equation}
\label{eqn:loss}
L(\Theta)=\frac{1}{n}\sum_{i=1}^n||F(\mathbf{Y}_i ; \Theta) - \mathbf{X}_i||^2,
\end{equation}
where $n$ is the number of training samples.
Using MSE as the loss function favors a high PSNR. The PSNR is a widely-used metric for quantitatively evaluating image restoration quality, and is at least partially related to the perceptual quality. It is worth noticing that the convolutional neural networks do not preclude the usage of other kinds of loss functions, if only the loss functions are derivable. If a better perceptually motivated metric is given during training, it is flexible for the network to adapt to that metric. On the contrary, such a flexibility is in general difficult to achieve for traditional ``hand-crafted'' methods.
Despite that the proposed model is trained favoring a high PSNR, we still observe satisfactory performance when the model is evaluated using alternative evaluation metrics, \eg~SSIM, MSSIM (see Section~\ref{sec:quality}).

The loss is minimized using stochastic gradient descent with the standard backpropagation~\cite{LeCun1998}. In particular, the weight matrices are updated as
\begin{equation}
\label{eqn:loss}
\Delta_{i+1}=0.9 \cdot \Delta_{i} - \eta \cdot \frac{\partial L}{\partial W_i^{\ell}},
\quad W_{i+1}^{\ell}=W_i^{\ell}+\Delta_{i+1},
\end{equation}
where $\ell \in \{1,2,3\}$ and $i$ are the indices of layers and iterations, $\eta$ is the learning rate, and $\frac{\partial L}{\partial W_i^{\ell}}$ is the derivative.
The filter weights of each layer are initialized by drawing randomly from a Gaussian distribution with zero mean and standard deviation 0.001 (and 0 for biases). The learning rate is $10^{-4}$ for the first two layers, and $10^{-5}$ for the last layer. We empirically find that a smaller learning rate in the last layer is important for the network to converge (similar to the denoising case \cite{jain2008natural}).

In the training phase, the ground truth images $\{\mathbf{X}_i\}$ are prepared as $f_{sub} \times f_{sub} \times c$-pixel sub-images randomly cropped from the training images. By ``sub-images'' we mean these samples are treated as small ``images'' rather than ``patches'', in the sense that ``patches'' are overlapping and require some averaging as post-processing but ``sub-images'' need not. To synthesize the low-resolution samples $\{\mathbf{Y}_i\}$, we blur a sub-image by a Gaussian kernel, sub-sample it by the upscaling factor, and upscale it by the same factor via bicubic interpolation.

To avoid border effects during training, all the convolutional layers have no padding, and the network produces a smaller output ($(f_{sub}-f_1-f_2-f_3+3)^2\times c$). The MSE loss function is evaluated only by the difference between the central pixels of $\mathbf{X}_i$ and the network output.
Although we use a fixed image size in training, the convolutional neural network can be applied on images of arbitrary sizes during testing.

We implement our model using the \emph{cuda-convnet} package \cite{Krizhevsky2012}.
We have also tried the \textit{Caffe} package~\cite{jia2014caffe} and observed similar performance.

\section{Experiments}
\label{sec:settings}

We first investigate the impact of using different datasets on the model performance. Next, we examine the filters learned by our approach. We then explore different architecture designs of the network, and study the relations between super-resolution performance and factors like depth, number of filters, and filter sizes. Subsequently, we compare our method with recent state-of-the-arts both quantitatively and qualitatively. Following \cite{Timofte2014}, super-resolution is only applied on the luminance channel (Y channel in YCbCr color space) in Sections~\ref{sec:ImageNet}-\ref{sec:results}, so $c=1$ in the first/last layer, and performance (\eg~PSNR and SSIM) is evaluated on the Y channel. At last, we extend the network to cope with color images and evaluate the performance on different channels.

\vspace{-0.2cm}

\subsection{Training Data}
\label{sec:ImageNet}

\begin{figure}[b]\tiny
\centering
  \includegraphics[width=\linewidth]{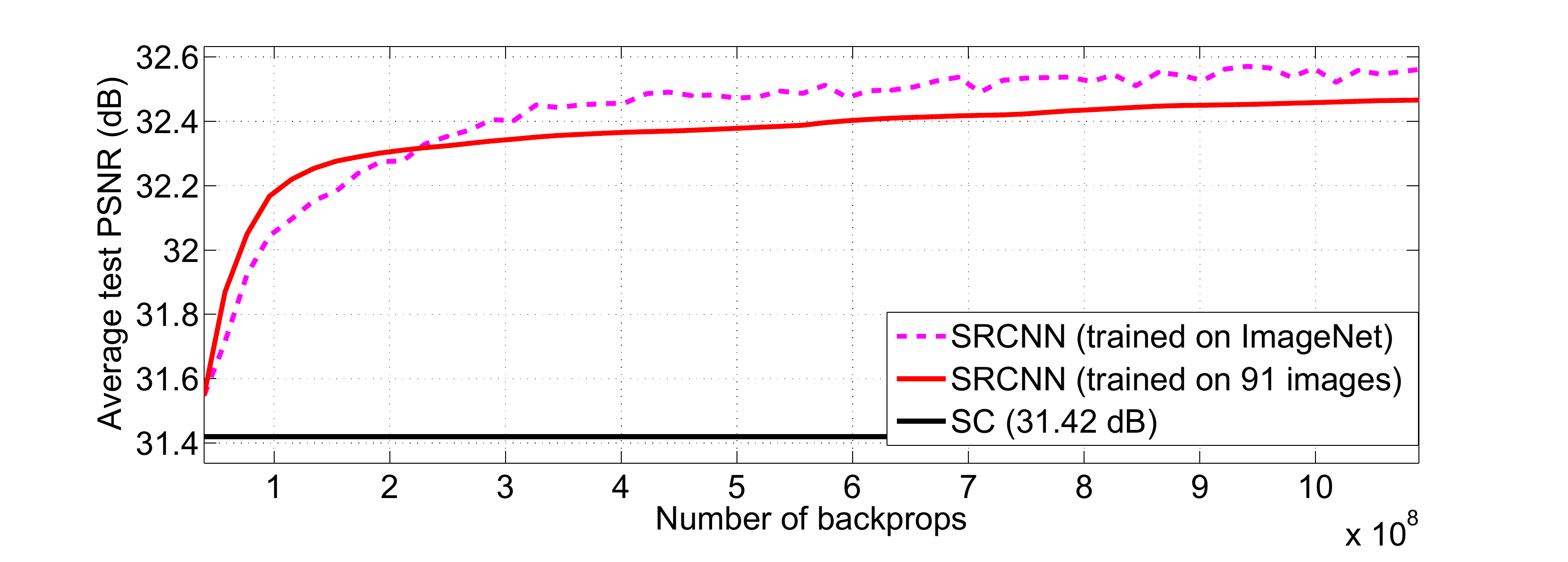}
\vskip -0.4cm
  \caption{Training with the much larger ImageNet dataset improves the performance over the use of 91 images.}\label{ImageNet}
  \vskip -0.4cm
\end{figure}

As shown in the literature, deep learning generally benefits from big data training.
For comparison, we use a relatively small training set~\cite{Yang2010a,Timofte2013} that consists of 91 images, and a large training set that consists of 395,909 images from the ILSVRC 2013 ImageNet detection training partition.
The size of training sub-images is $f_{sub}=33$. Thus the 91-image dataset can be decomposed into 24,800 sub-images, which are extracted from original images with a stride of 14. Whereas the ImageNet provides over 5 million sub-images even using a stride of 33. We use the basic network settings, \ie~$f_1=9$, $f_2=1$, $f_3=5$, $n_1=64$, and $n_2=32$.
We use the \textit{Set5}~\cite{Bevilacqua2012} as the validation set.
We observe a similar trend even if we use the larger \textit{Set14} set~\cite{Zeyde2012}.
The upscaling factor is 3.
We use the sparse-coding-based method~\cite{Yang2010a} as our baseline, which achieves an average PSNR value of 31.42 dB.

The test convergence curves of using different training sets are shown in Figure~\ref{ImageNet}.
The training time on ImageNet is about the same as on the 91-image dataset since the number of backpropagations is the same.
As can be observed, with the same number of backpropagations (\ie $8\times 10^{8}$), the SRCNN$+$ImageNet achieves \textbf{32.52 dB}, higher than 32.39 dB yielded by that trained on 91 images. The results positively indicate that SRCNN performance may be further boosted using a larger training set, but the effect of big data is not as impressive as that shown in high-level vision problems~\cite{Krizhevsky2012}. This is mainly because that the 91 images have already captured sufficient variability of natural images. On the other hand, our SRCNN is a relatively small network (8,032 parameters), which could not overfit the 91 images (24,800 samples). Nevertheless, we adopt the ImageNet, which contains more diverse data, as the default training set in the following experiments.

\subsection{Learned Filters for Super-Resolution}
\label{sec:learned_filter}

\begin{figure}[t]
\centering
  \includegraphics[width=0.9\linewidth]{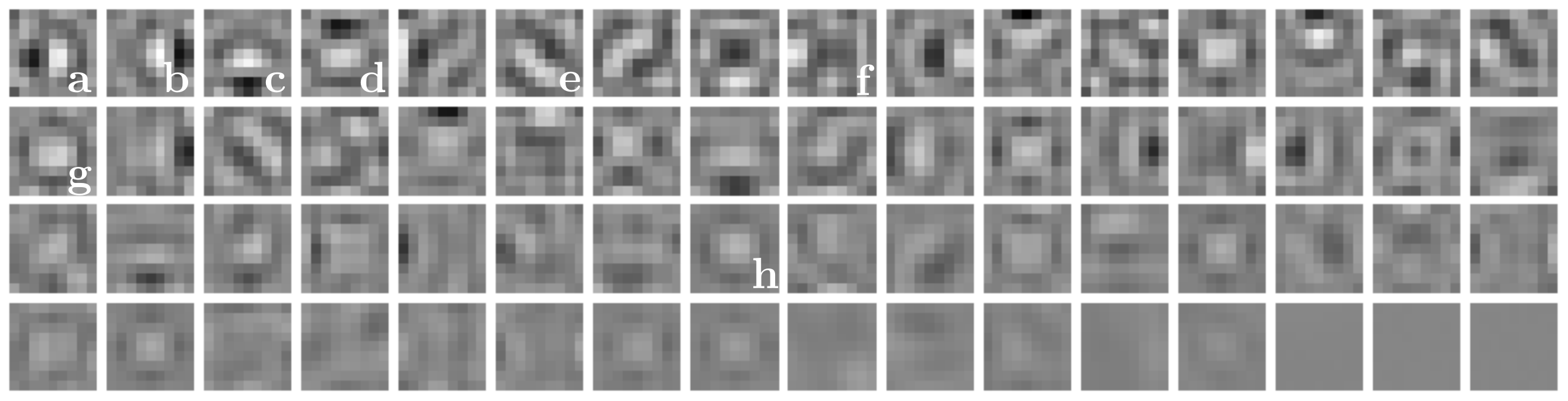}
\vskip -0.25cm
\caption{The figure shows the first-layer filters trained on ImageNet with an upscaling factor 3. The filters are organized based on their respective variances.}
  \label{fig:patterns}
\vskip -0.4cm
\end{figure}

\begin{figure}[t]
\centering
  \includegraphics[width=0.9\linewidth]{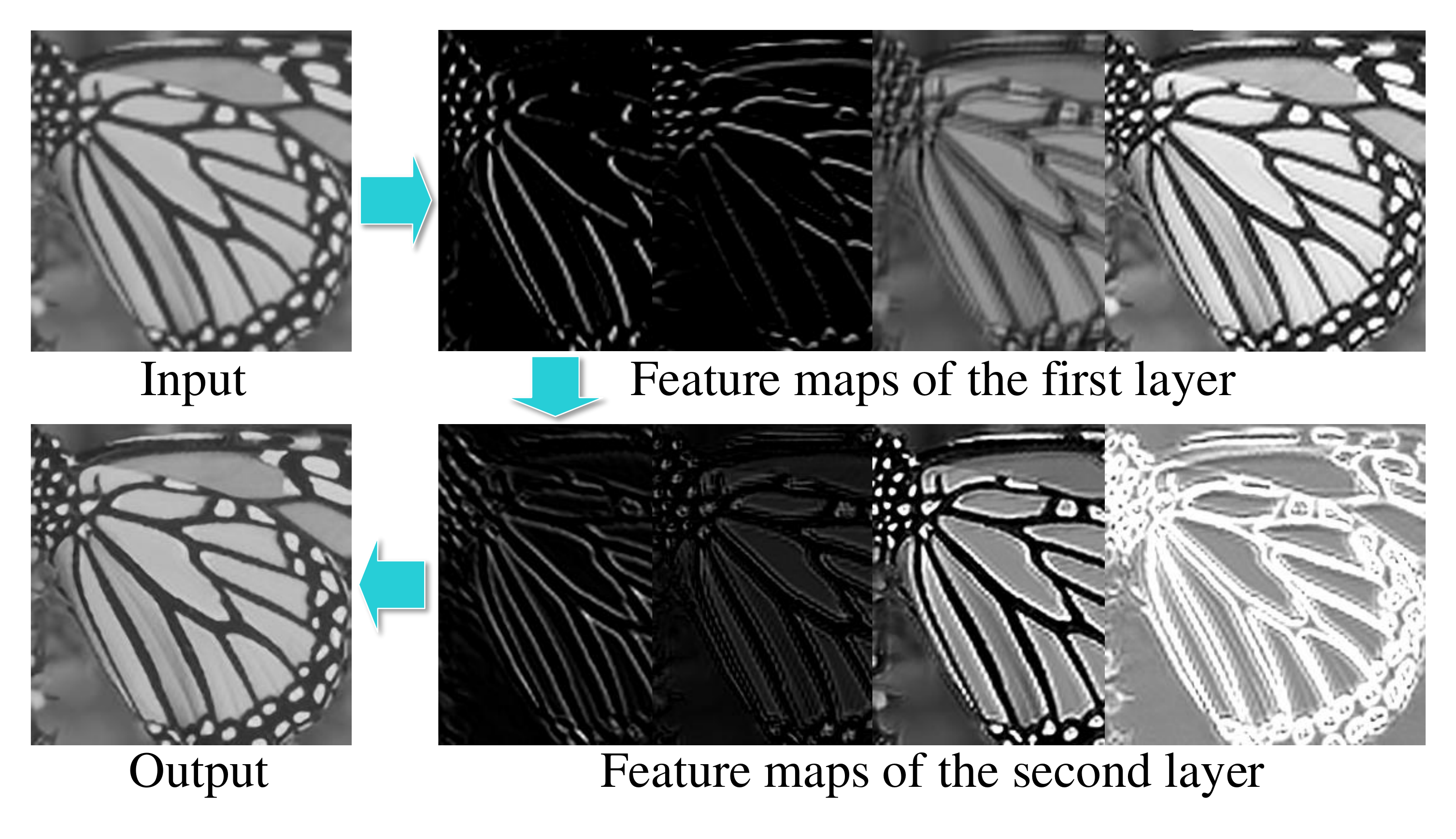}
\vskip -0.25cm
\caption{Example feature maps of different layers.}
  \label{fig:featuremap}
\vskip -0.4cm
\end{figure}

Figure~\ref{fig:patterns} shows examples of learned first-layer filters trained on the ImageNet by an upscaling factor 3. Please refer to our published implementation for upscaling factors 2 and 4.
Interestingly, each learned filter has its specific functionality.
For instance, the filters $g$ and $h$ are like Laplacian/Gaussian filters, the filters $a$ - $e$ are like edge detectors at different directions, and the filter $f$ is like a texture extractor.
Example feature maps of different layers are shown in figure~\ref{fig:featuremap}. Obviously, feature maps of the first layer contain different structures (\eg~edges at different directions), while that of the second layer are mainly different on intensities.

\subsection{Model and Performance Trade-offs}
\label{sec:model_variants}

Based on the basic network settings (\ie~$f_1=9$, $f_2=1$, $f_3=5$, $n_1=64$, and $n_2=32$),
we will progressively modify some of these parameters to investigate the best trade-off between performance and speed, and study the relations between performance and parameters.

\subsubsection{Filter number}
\label{sec:Filternum}

In general, the performance would improve if we increase the network width\footnote{We use `width' to term the number of filters in a layer, following~\cite{he2014convolutional}. The term `width' may have other meanings in the literature.}, \ie~adding more filters, at the cost of running time. Specifically, based on our network default settings of $n_1=64$ and $n_2=32$, we conduct two experiments: (i) one is with a larger network with $n_1=128$ and $n_2=64$, and (ii) the other is with a smaller network with $n_1=32$ and $n_2=16$.
Similar to Section~\ref{sec:ImageNet}, we also train the two models on ImageNet and test on Set5 with an upscaling factor 3. The results observed at $8\times10^8$ backpropagations are shown in Table~\ref{tab:filter_number}. It is clear that superior performance could be achieved by increasing the width. However, if a fast restoration speed is desired, a small network width is preferred, which could still achieve better performance than the sparse-coding-based method (31.42 dB).

\begin{table}[h]\footnotesize
\caption{The results of using different filter numbers in SRCNN. Training is performed on ImageNet whilst the evaluation is conducted on the Set5 dataset.}\label{tab:filter_number}
\vspace{-0.4cm}
\begin{center}
\begin{tabular}{|c|c|c|c|c|c|}
\hline
\multicolumn{2}{|c|}{$n_1=128$} &  \multicolumn{2}{c|}{$n_1=64$} &  \multicolumn{2}{c|}{$n_1=32$} \\
\multicolumn{2}{|c|}{$n_2=64$} &  \multicolumn{2}{c|}{$n_2=32$} &  \multicolumn{2}{c|}{$n_2=16$} \\
\hline
PSNR&Time (sec) & PSNR&Time (sec)& PSNR&Time (sec) \\

\hline\hline
32.60 & 0.60 & 32.52 & 0.18 & 32.26 & 0.05 \\

\hline
\end{tabular}
\end{center}
\vspace{-0.45cm}
\end{table}

\subsubsection{Filter size}
\label{sec:Filtersize}
\begin{figure}\tiny
\centering
  \includegraphics[width=\linewidth]{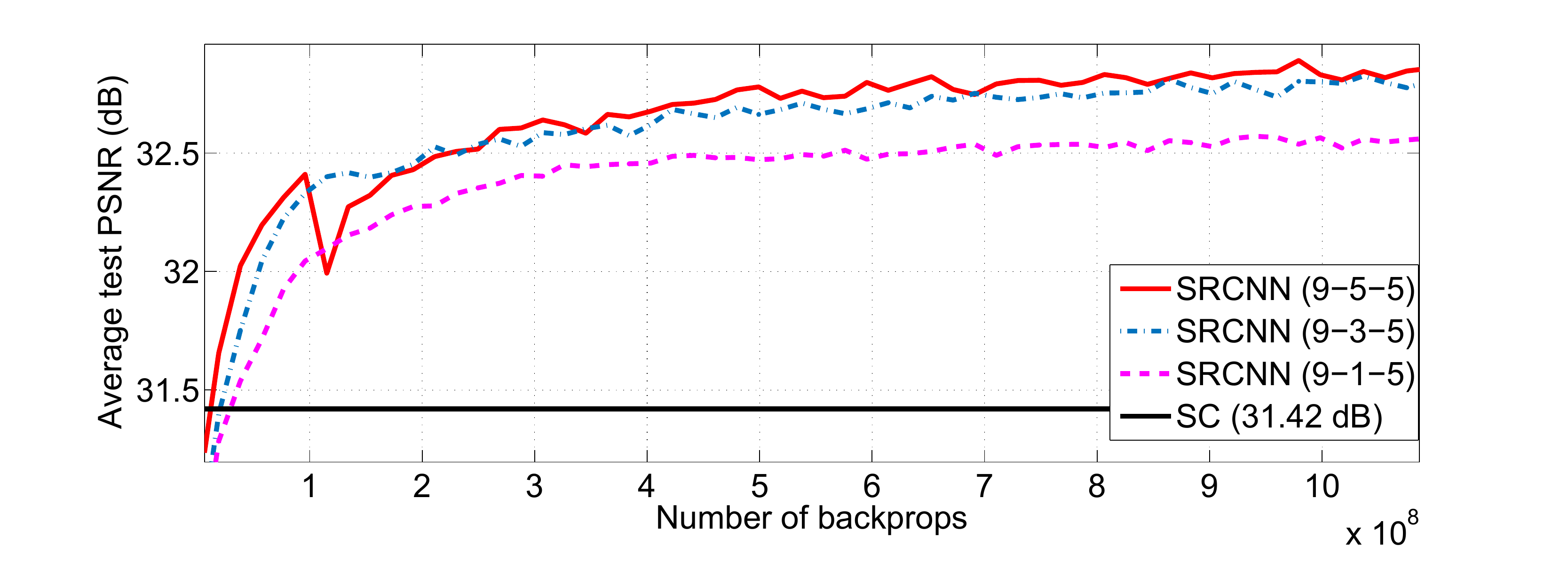}
\vskip -0.4cm
  \caption{A larger filter size leads to better results.}\label{wider_net}
\vskip -0.4cm
\end{figure}

In this section, we examine the network sensitivity to different filter sizes.
In previous experiments, we set filter size $f_1=9$, $f_2=1$ and $f_3=5$, and the network could be denoted as 9-1-5.
First, to be consistent with sparse-coding-based methods, we fix the filter size of the second layer to be $f_2=1$, and enlarge the filter size of other layers to $f_1=11$ and $f_3=7$ (11-1-7).
All the other settings remain the same with Section~\ref{sec:ImageNet}. The results with an upscaling factor 3 on Set5 are 32.57 dB, which is slightly higher than the 32.52 dB reported in Section~\ref{sec:ImageNet}.
This indicates that a reasonably larger filter size could grasp richer structural information, which in turn lead to better results.

Then we further examine networks with a larger filter size of the second layer. Specifically, we fix the filter size $f_1=9$, $f_3=5$, and enlarge the filter size of the second layer to be (i) $f_2=3$ (9-3-5) and (ii) $f_2=5$ (9-5-5). Convergence curves in Figure~\ref{wider_net} show that using a larger filter size could significantly improve the performance. Specifically, the average PSNR values achieved by 9-3-5 and 9-5-5 on Set5 with $8\times10^8$ backpropagations are 32.66 dB and 32.75 dB, respectively. The results suggest that utilizing neighborhood information in the mapping stage is beneficial.

However, the deployment speed will also decrease with a larger filter size. For example, the number of parameters of 9-1-5, 9-3-5, and 9-5-5 is 8,032, 24,416, and 57,184 respectively. The complexity of 9-5-5 is almost twice of  9-3-5, but the performance improvement is marginal. Therefore, the choice of the network scale should always be a trade-off between performance and speed.

\subsubsection{Number of layers}
\label{sec:Deeper}

\begin{figure}[t]\tiny
\centering
\subfigure[9-1-5 vs. 9-1-1-5]{
  \label{deep1}
  \includegraphics[width=\linewidth]{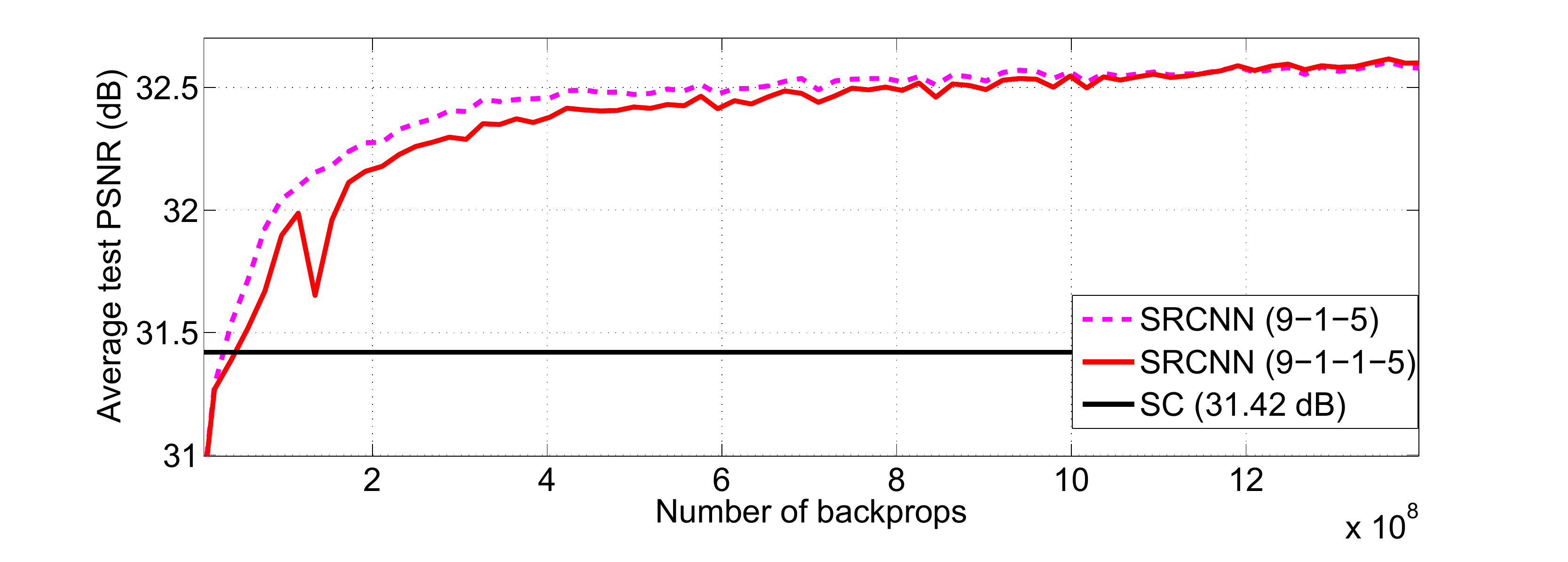}
}
\vskip -0.2cm
\subfigure[9-3-5 vs. 9-3-1-5]{
  \label{deep2}
  \includegraphics[width=\linewidth]{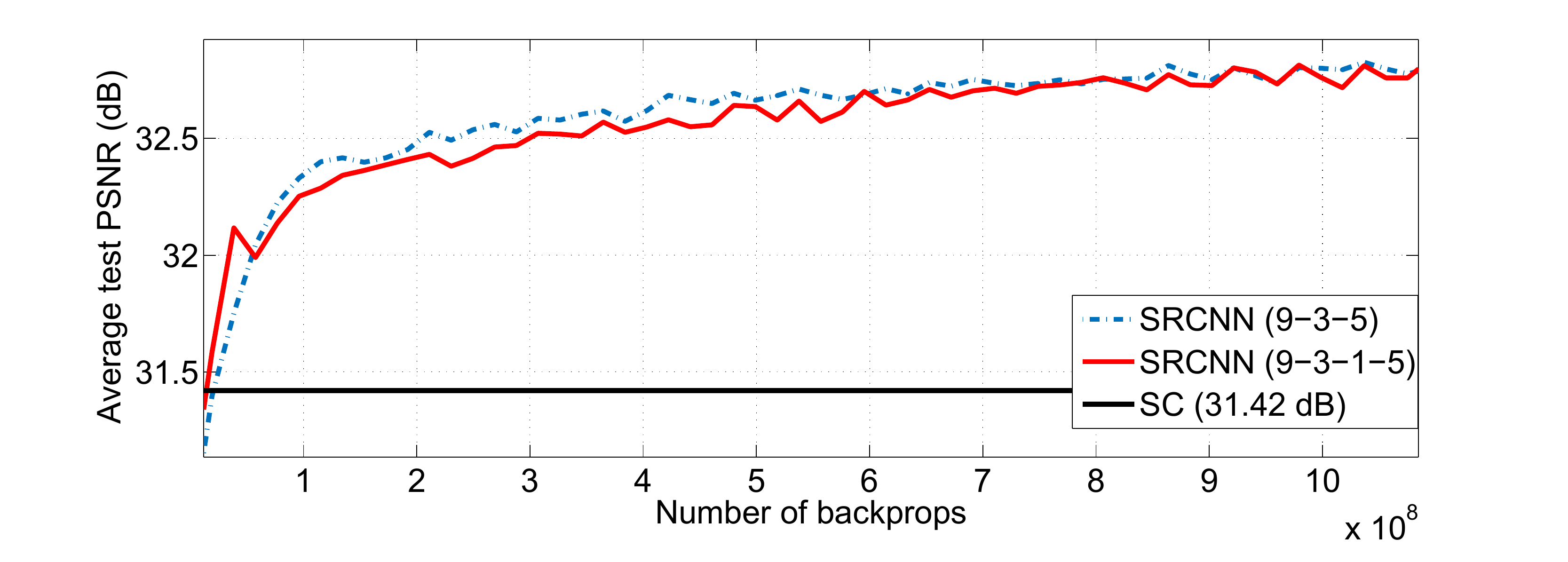}
}
\vskip -0.2cm
\subfigure[9-5-5 vs. 9-5-1-5]{
  \label{deep3}
  \includegraphics[width=\linewidth]{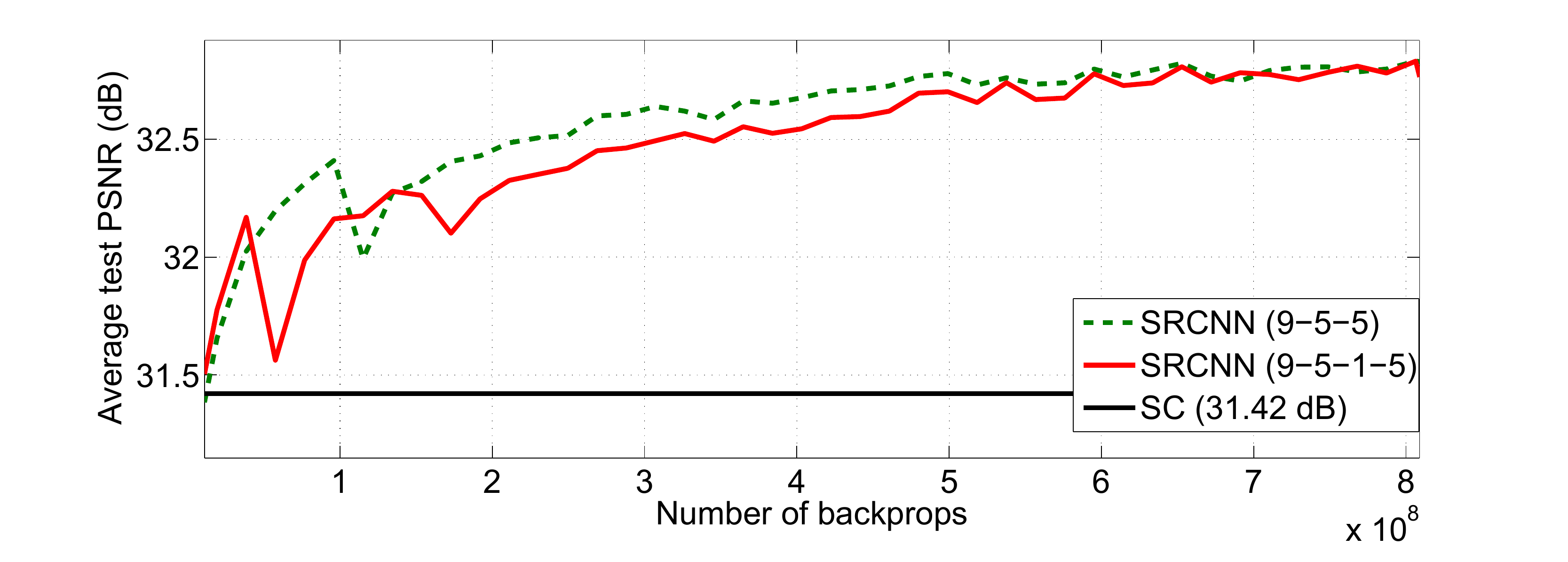}
}
\vskip -0.25cm
  \caption{Comparisons between three-layer and four-layer networks.}
\vskip -0.4cm
\end{figure}

Recent study by He and Sun~\cite{he2014convolutional} suggests that CNN could benefit from increasing the depth of network moderately. Here, we try deeper structures by adding another non-linear mapping layer, which has $n_{22}=16$ filters with size $f_{22}=1$. We conduct three controlled experiments, \ie~9-1-1-5, 9-3-1-5, 9-5-1-5, which add an additional layer on 9-1-5, 9-3-5, and 9-5-5, respectively. The initialization scheme and learning rate of the additional layer are the same as the second layer. From Figures~\ref{deep1},~\ref{deep2} and~\ref{deep3}, we can observe that the four-layer networks converge slower than the three-layer network. Nevertheless, given enough training time, the deeper networks will finally catch up and converge to the three-layer ones.

\begin{figure}[t]\tiny
\centering

\subfigure[9-1-1-5 ($n_{22}=32$) and 9-1-1-1-5 ($n_{22}=32, n_{23}=16$)]{
  \label{deep4}
  \includegraphics[width=\linewidth]{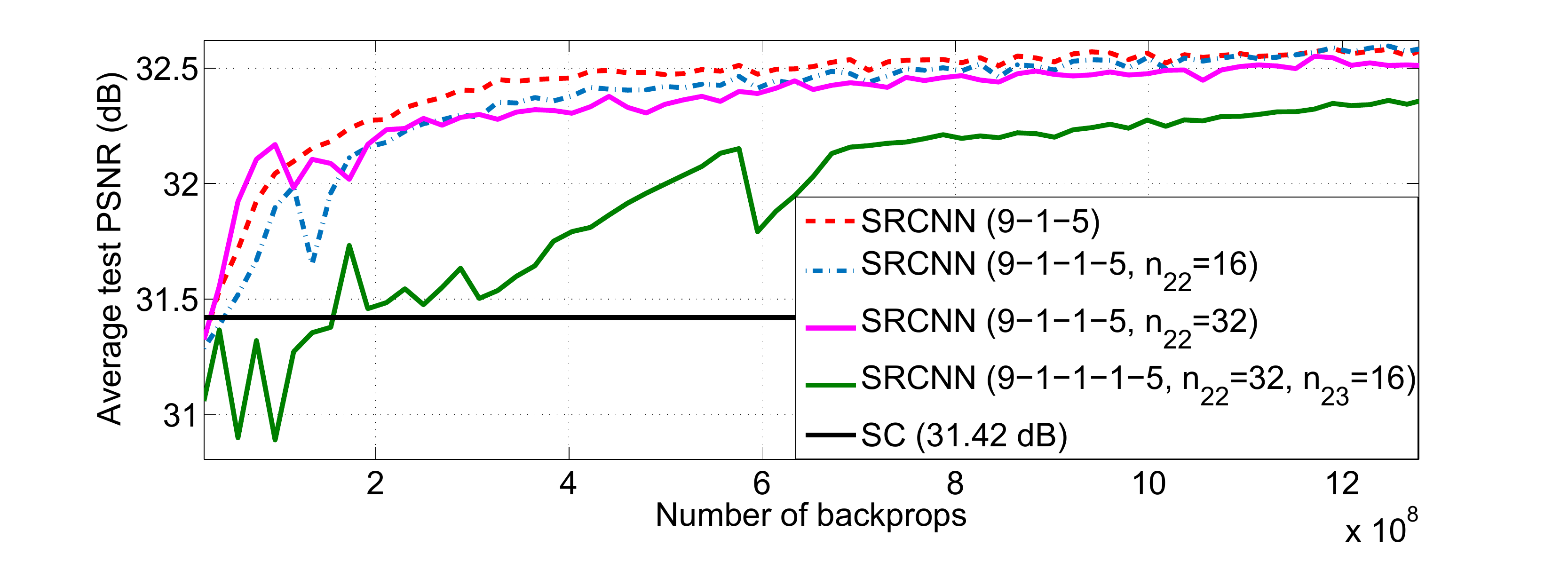}
}
\vskip -0.15cm
\subfigure[9-3-3-5 and 9-3-3-3]{
  \label{deep5}
  \includegraphics[width=\linewidth]{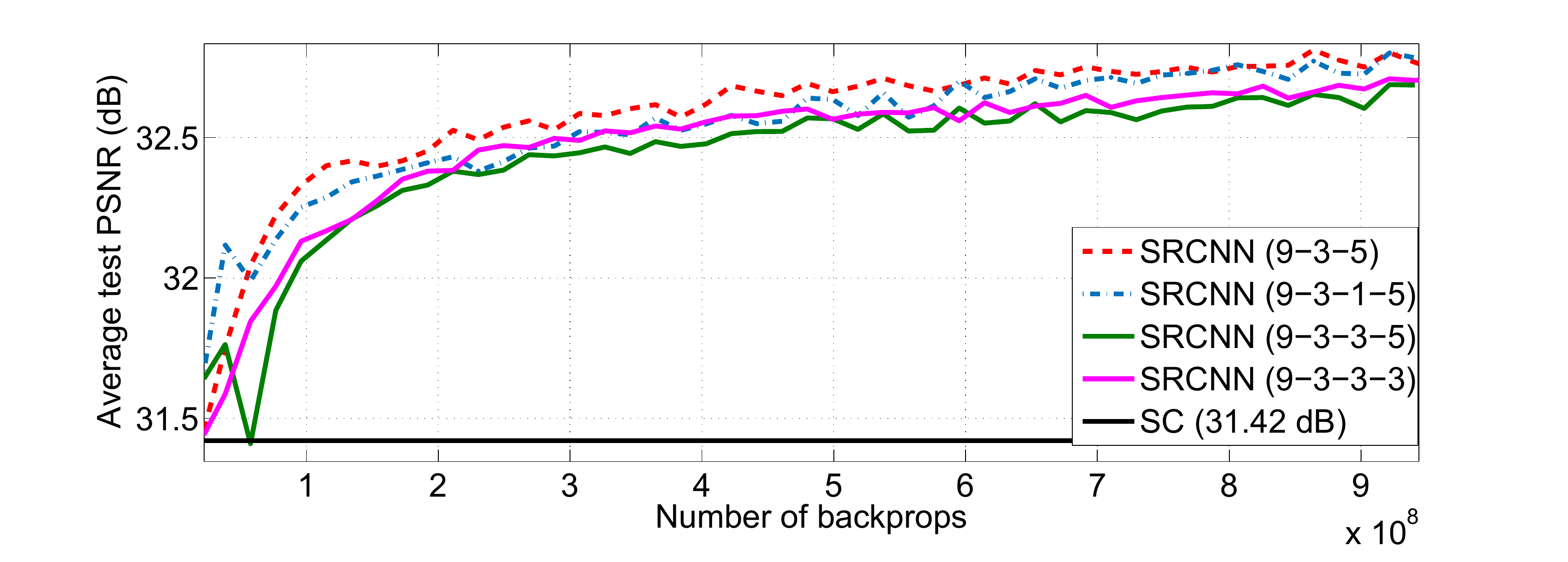}
}
\vskip -0.25cm
  \caption{Deeper structure does not always lead to better results.}
\vskip -0.4cm
\end{figure}

The effectiveness of deeper structures for super resolution is found not as apparent as that shown in image classification~\cite{he2014convolutional}.
Furthermore, we find that deeper networks do not always result in better performance. Specifically, if we add an additional layer with $n_{22}=32$ filters on 9-1-5 network, then the performance degrades and fails to surpass the three-layer network (see Figure~\ref{deep4}). If we go deeper by adding two non-linear mapping layers with $n_{22}=32$ and $n_{23}=16$ filters on 9-1-5, then we have to set a smaller learning rate to ensure convergence, but we still do not observe superior performance after a week of training (see Figure~\ref{deep4}). We also tried to enlarge the filter size of the additional layer to $f_{22}=3$, and explore two deep structures -- 9-3-3-5 and 9-3-3-3. However, from the convergence curves shown in Figure~\ref{deep5}, these two networks do not show better results than the 9-3-1-5 network.

All these experiments indicate that it is not ``the deeper the better'' in this deep model for super-resolution. It may be caused by the difficulty of training. Our CNN network contains no pooling layer or full-connected layer, thus it is sensitive to the initialization parameters and learning rate. When we go deeper (\eg~4 or 5 layers), we find it hard to set appropriate learning rates that guarantee convergence. Even it converges, the network may fall into a bad local minimum, and the learned filters are of less diversity even given enough training time. This phenomenon is also observed in [16], where improper increase of depth leads to accuracy saturation or degradation for image classification. Why ``deeper is not better'' is still an open question, which requires investigations to better understand gradients and training dynamics in deep architectures.
Therefore, we still adopt three-layer networks in the following experiments.

\subsection{Comparisons to State-of-the-Arts}
\label{sec:results}

In this section, we show the quantitative and qualitative results of our method in comparison to state-of-the-art methods.
We adopt the model with good performance-speed trade-off: a three-layer network with $f_1=9$, $f_2=5$, $f_3=5$, $n_1=64$, and $n_2=32$ trained on the ImageNet. For each upscaling factor $\in\left\{2,3,4\right\}$, we train a specific network for that factor\footnote{In the area of denoising \cite{Burger2012}, for each noise level a specific network is trained.}.

\noindent
\textbf{Comparisons.}
We compare our \textit{SRCNN} with the state-of-the-art SR methods:
\begin{packed_itemize}
\item \textit{SC} - sparse coding-based method of Yang~\etal~\cite{Yang2010a}
\item \textit{NE+LLE} - neighbour embedding + locally linear embedding method~\cite{Chang2004}
\item \textit{ANR} - Anchored Neighbourhood Regression method~\cite{Timofte2013}
\item \textit{A+} - Adjusted Anchored Neighbourhood Regression method~\cite{Timofte2014}, and
\item \textit{KK} - the method described in~\cite{Kim2010}, which achieves the best performance among external example-based methods, according to the comprehensive evaluation conducted in Yang~\etal's work~\cite{Yang2014}
\end{packed_itemize}
The implementations are all from the publicly available codes provided by the authors, and all images are down-sampled using the same bicubic kernel.

\noindent
\textbf{Test set.}
The \textit{Set5}~\cite{Bevilacqua2012} (5 images), \textit{Set14}~\cite{Zeyde2012} (14 images) and \textit{BSD200}~\cite{martin2001database} (200 images)\footnote{We use the same 200 images as in \cite{Yang2014}.} are used to evaluate the performance of upscaling factors 2, 3, and 4.

\noindent
\textbf{Evaluation metrics.}
Apart from the widely used PSNR and SSIM~\cite{wang2004image} indices, we also adopt another four evaluation matrices, namely information fidelity criterion (IFC)~\cite{sheikh2005information}, noise quality measure (NQM)~\cite{damera2000image}, weighted peak signal-to-noise ratio (WPSNR) and multi-scale structure similarity index (MSSSIM)~\cite{wang2003multiscale}, which obtain high correlation with the human perceptual scores as reported in~\cite{Yang2014}.

\subsubsection{Quantitative and qualitative evaluation}
\label{sec:quality}

As shown in Tables~\ref{set5},~\ref{set14} and~\ref{BSD200}, the proposed SRCNN yields the highest scores in most evaluation matrices in all experiments\footnote{The PSNR value of each image can be found in the supplementary file.}. Note that our SRCNN results are based on the checkpoint of $8\times 10^{8}$ backpropagations.
Specifically, for the upscaling factor 3, the average gains on PSNR achieved by SRCNN are 0.15 dB, 0.17 dB, and 0.13 dB, higher than the next best approach, A+~\cite{Timofte2014}, on the three datasets.
When we take a look at other evaluation metrics, we observe that SC, to our surprise, gets even lower scores than the bicubic interpolation on IFC and NQM. It is clear that the results of SC are more visually pleasing than that of bicubic interpolation. This indicates that these two metrics may not truthfully reveal the image quality. Thus, regardless of these two metrics, SRCNN achieves the best performance among all methods and scaling factors.

\begin{figure}\tiny
\centering
  \includegraphics[width=\linewidth]{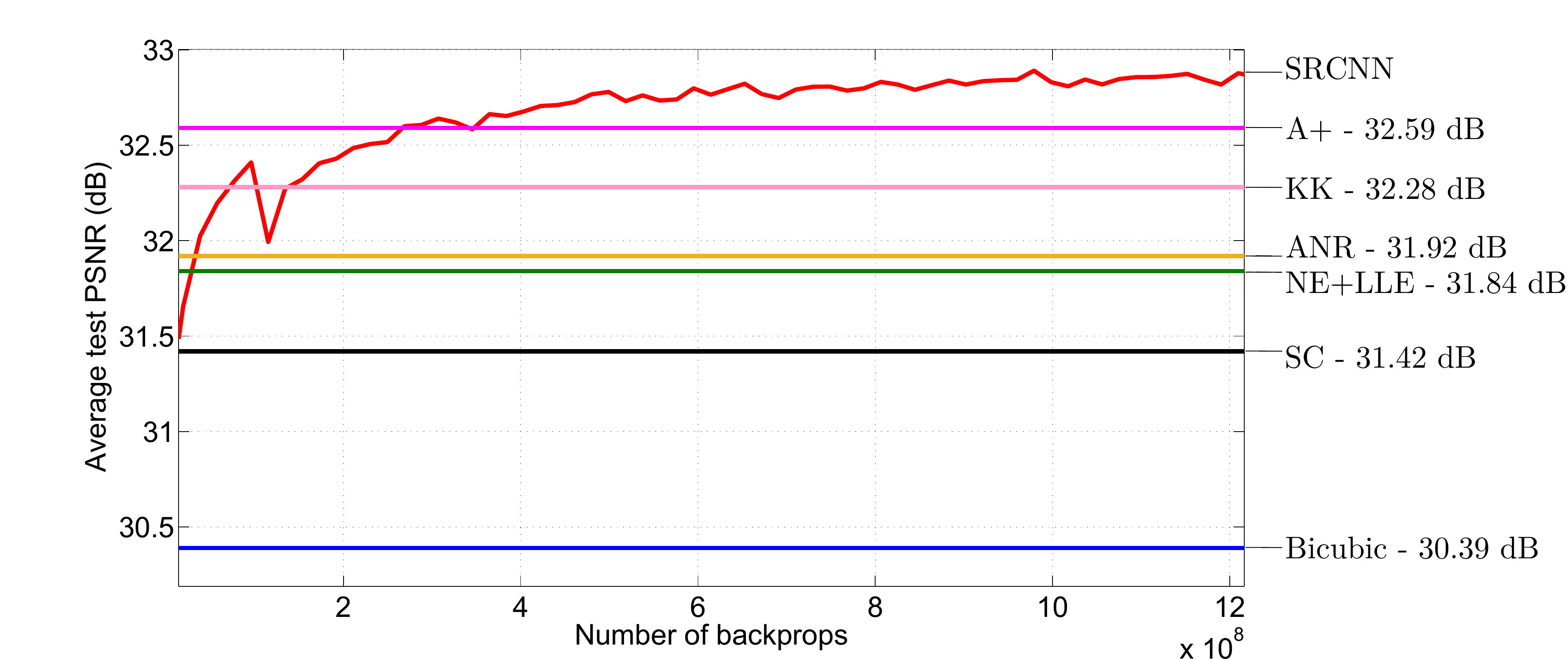}
\vskip -0.25cm
  \caption{The test convergence curve of SRCNN and results of other methods on the Set5 dataset.}\label{fig:comparison}
  \vskip -0.45cm
\end{figure}

It is worth pointing out that SRCNN surpasses the bicubic baseline at the very beginning of the learning stage (see Figure~\ref{fig:overview}), and with moderate training, SRCNN outperforms existing state-of-the-art methods (see Figure~\ref{ImageNet}). Yet, the performance is far from converge. We conjecture that better results can be obtained given longer training time (see Figure~\ref{fig:comparison}).

\begin{figure*}[p]\small
\begin{center}
\includegraphics[width=0.8\linewidth]{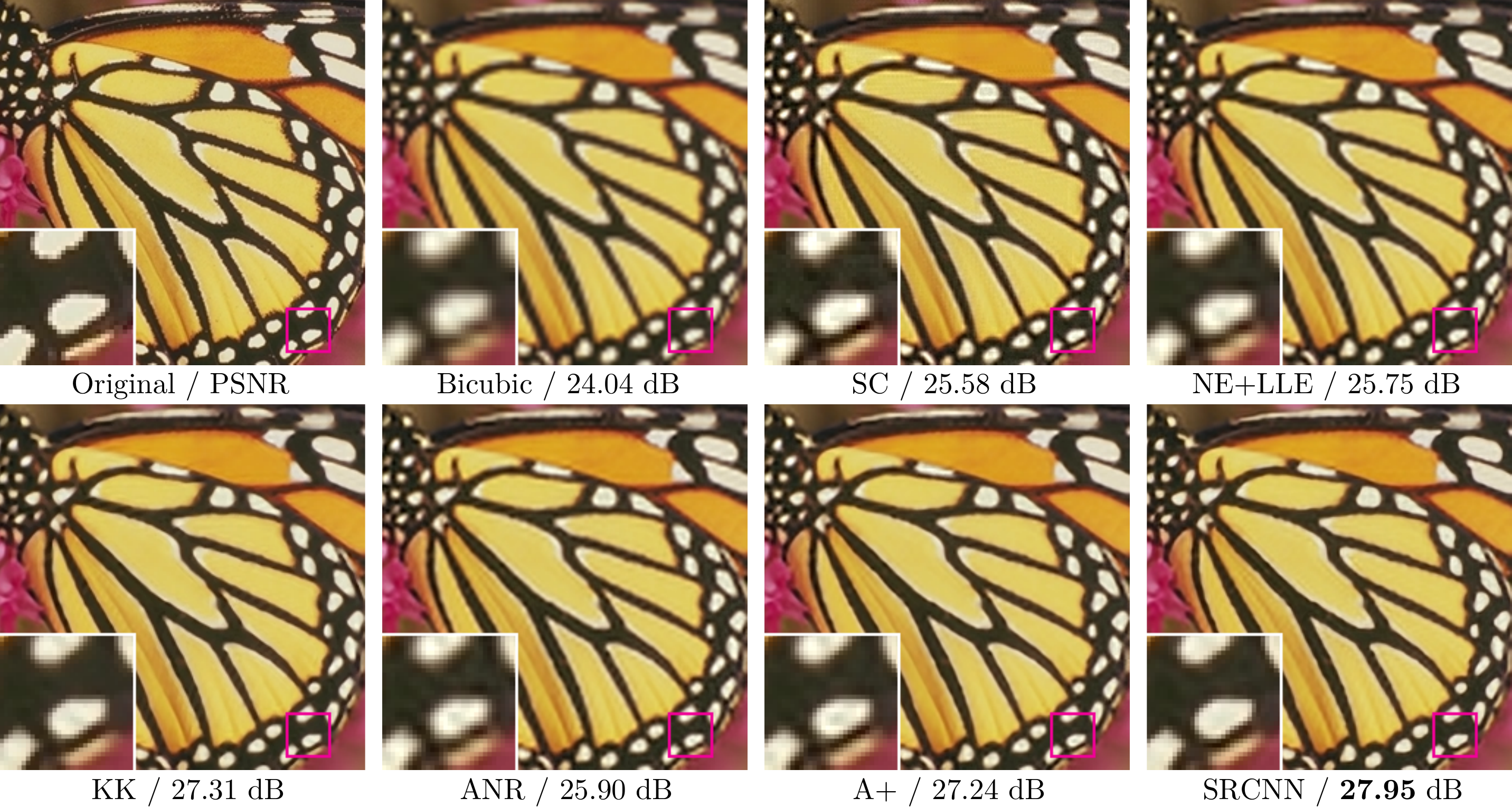}
\vskip -0.25cm
\caption{The ``butterfly" image from Set5 with an upscaling factor 3.}
\label{fig:butterfly}
\vspace{-0.5cm}
\end{center}
\end{figure*}

\begin{figure*}[p]\small
\begin{center}
\includegraphics[width=0.8\linewidth]{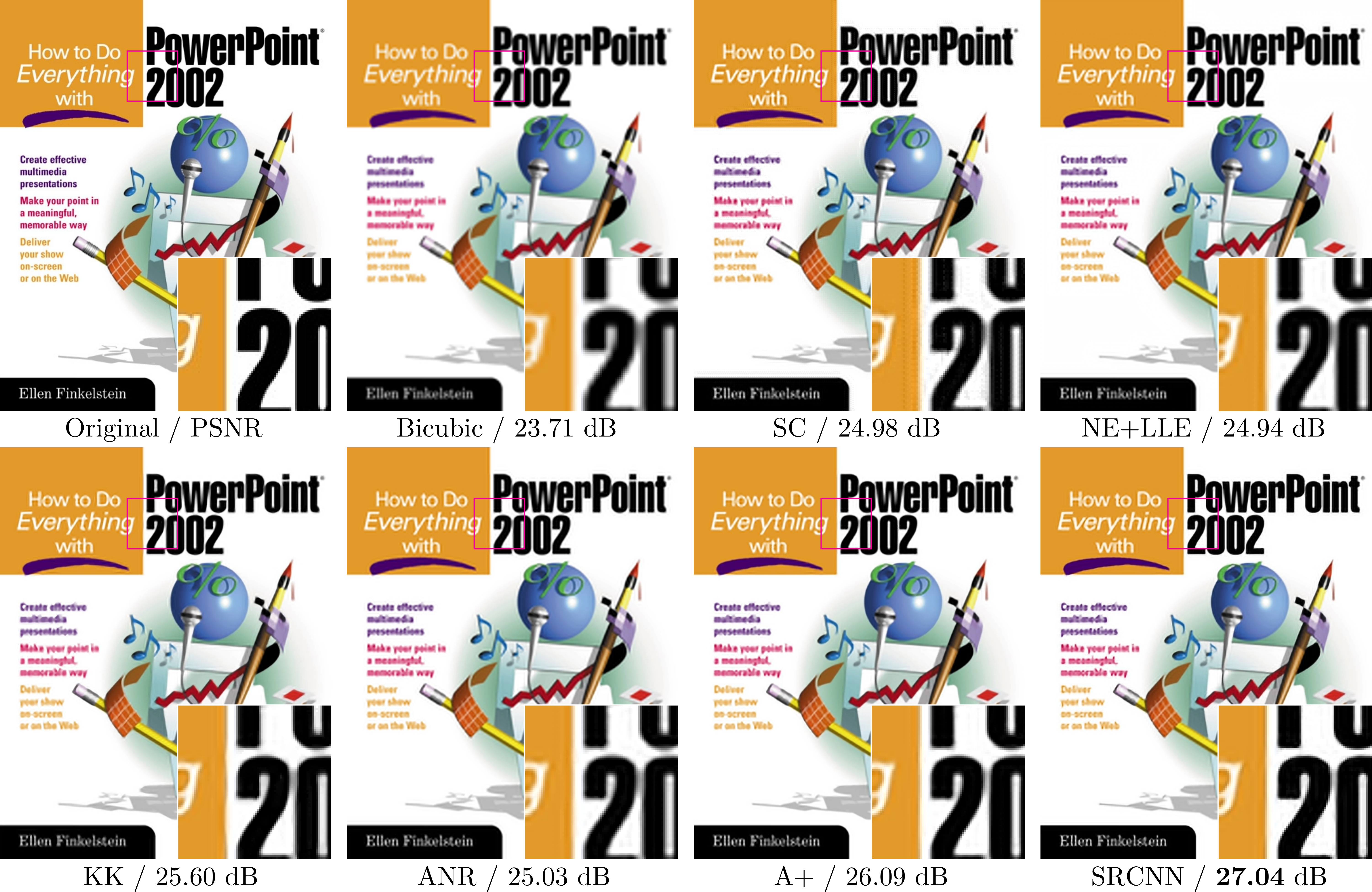}
\vskip -0.25cm
\caption{The ``ppt3" image from Set14 with an upscaling factor 3.}
\label{fig:ppt}
\vspace{-0.5cm}
\end{center}
\end{figure*}

\begin{figure*}[p]\small
\begin{center}
\includegraphics[width=0.8\linewidth]{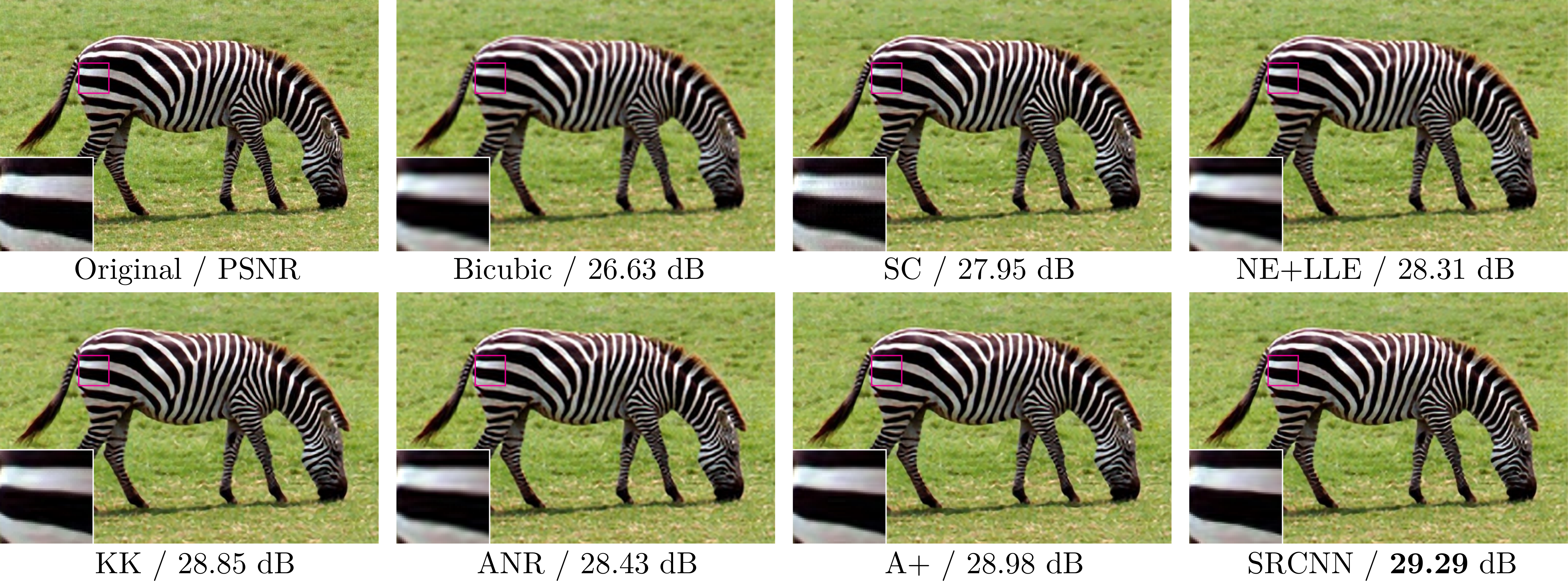}
\vskip -0.25cm
\caption{The ``zebra" image from Set14 with an upscaling factor 3.}
\label{fig:zebra}
\vspace{-0.5cm}
\end{center}
\end{figure*}

Figures~\ref{fig:butterfly},~\ref{fig:ppt} and~\ref{fig:zebra} show the super-resolution results of different approaches by an upscaling factor 3. As can be observed, the SRCNN produces much sharper edges than other approaches without any obvious artifacts across the image.

\begin{figure}[b]
\centering
  \includegraphics[width=\linewidth]{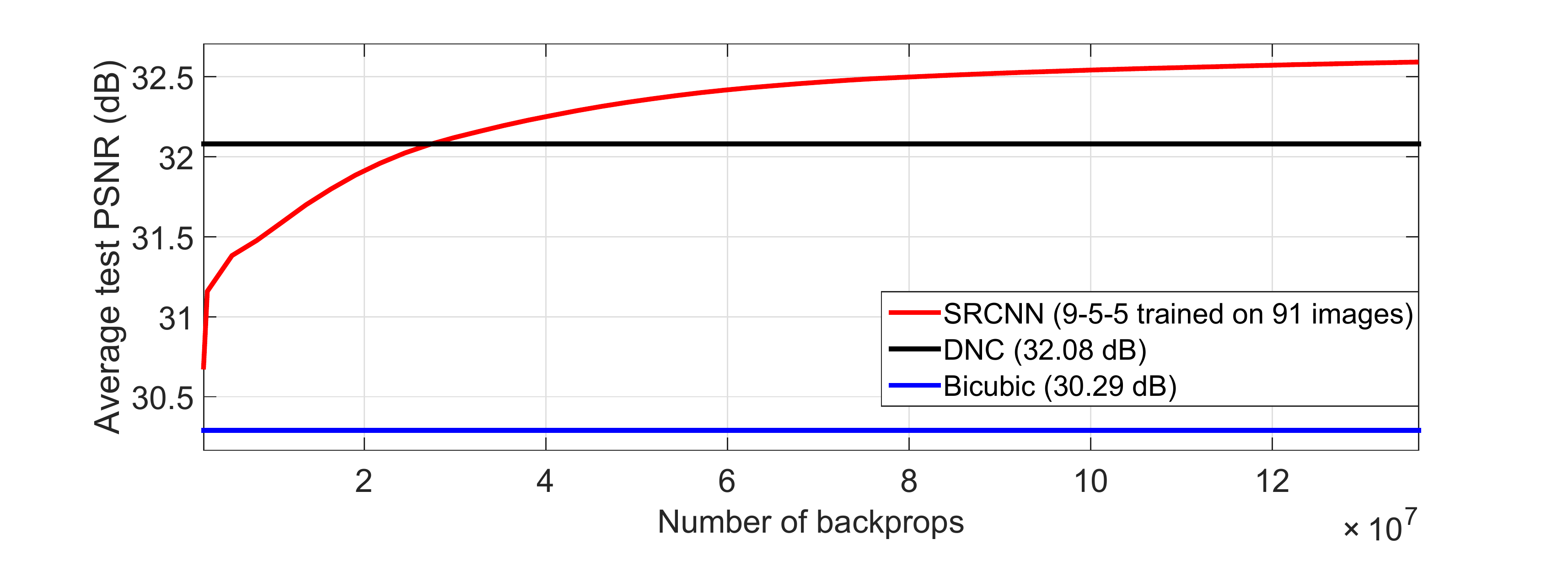}
\vskip -0.25cm
  \caption{The test convergence curve of SRCNN and the result of DNC on the Set5 dataset.}
  \label{fig:DNC}
  \vskip -0.45cm
\end{figure}

In addition, we report to another recent deep learning method for image super-resolution (DNC) of Cui~\etal~\cite{Cui2014}. As they employ a different blur kernel (a Gaussian filter with a standard deviation of 0.55), we train a specific network (9-5-5) using the same blur kernel as DNC for fair quantitative comparison. The upscaling factor is 3 and the training set is the 91-image dataset. From the convergence curve shown in Figure~\ref{fig:DNC}, we observe that our SRCNN surpasses DNC with just $2.7\times 10^7$ backprops, and a larger margin can be obtained given longer training time. This also demonstrates that the end-to-end learning is superior to DNC, even if that model is already ``deep''.


\begin{table*}[p]\footnotesize
\caption{The average results of PSNR (dB), SSIM, IFC, NQM, WPSNR (dB) and MSSIM on the Set5 dataset.}\label{set5}
\vskip -0.35cm
\begin{center}
\begin{tabular}{|c|c|c|c|c|c|c|c|c|}
\hline
Eval. Mat & Scale& Bicubic & SC~\cite{Yang2010a} &NE+LLE~\cite{Chang2004} & KK~\cite{Kim2010} & ANR~\cite{Timofte2013} & A+~\cite{Timofte2013} & SRCNN\\
\hline\hline
     & 2 & 33.66 & - & 35.77 & 36.20 & 35.83 & 36.54 & \textbf{36.66} \\
PSNR & 3 & 30.39 & 31.42 & 31.84 & 32.28 & 31.92 & 32.59 & \textbf{32.75}  \\
     & 4 & 28.42 & -& 29.61 & 30.03 & 29.69 & 30.28 & \textbf{30.49} \\
\hline\hline
     & 2 & 0.9299 & -& 0.9490 & 0.9511 & 0.9499 & \textbf{0.9544} & 0.9542 \\
SSIM & 3 & 0.8682 & 0.8821 & 0.8956 & 0.9033 & 0.8968 & 0.9088 & \textbf{0.9090} \\
     & 4 & 0.8104 & -& 0.8402 & 0.8541 & 0.8419 & 0.8603 & \textbf{0.8628} \\
\hline\hline
     & 2 & 6.10 & -& 7.84 & 6.87 & 8.09 & \textbf{8.48} & 8.05 \\
IFC  & 3 & 3.52 & 3.16 & 4.40 & 4.14 & 4.52 & \textbf{4.84} & 4.58 \\
     & 4 & 2.35 & -& 2.94 & 2.81 & 3.02 & \textbf{3.26} & 3.01  \\
\hline\hline
     & 2 & 36.73 & -& 42.90 & 39.49 & 43.28 & \textbf{44.58} & 41.13 \\
NQM  & 3 & 27.54 & 27.29 & 32.77 & 32.10 & 33.10 & \textbf{34.48} & 33.21 \\
     & 4 & 21.42 & -& 25.56 & 24.99 & 25.72 & \textbf{26.97} & 25.96 \\
\hline\hline
     & 2 & 50.06 & -& 58.45 & 57.15 & 58.61 & \textbf{60.06} & 59.49 \\
WPSNR  & 3 & 41.65 & 43.64 & 45.81 & 46.22 & 46.02 & \textbf{47.17} & 47.10 \\
     & 4 & 37.21 & -& 39.85 & 40.40 & 40.01 & 41.03 & \textbf{41.13} \\
\hline\hline
     & 2 & 0.9915 & -& 0.9953 & 0.9953 & 0.9954 & \textbf{0.9960} & 0.9959 \\
MSSSIM  & 3 & 0.9754 & 0.9797 & 0.9841 & 0.9853 & 0.9844 & \textbf{0.9867} & 0.9866\\
     & 4 & 0.9516 & -& 0.9666 & 0.9695 & 0.9672 & 0.9720 & \textbf{0.9725} \\
\hline
\end{tabular}
\end{center}
\vspace{-0.25cm}
\end{table*}

\begin{table*}[p]\footnotesize
\caption{The average results of PSNR (dB), SSIM, IFC, NQM, WPSNR (dB) and MSSIM on the Set14 dataset.}\label{set14}
\vskip -0.35cm
\begin{center}
\begin{tabular}{|c|c|c|c|c|c|c|c|c|}
\hline
Eval. Mat & Scale& Bicubic & SC~\cite{Yang2010a} &NE+LLE~\cite{Chang2004} & KK~\cite{Kim2010} & ANR~\cite{Timofte2013} & A+~\cite{Timofte2013} & SRCNN\\
\hline\hline
     & 2 & 30.23 & - & 31.76 & 32.11 & 31.80 & 32.28 & \textbf{32.45} \\
PSNR & 3 & 27.54 & 28.31 & 28.60 & 28.94 & 28.65 & 29.13 & \textbf{29.30}  \\
     & 4 & 26.00 & -& 26.81 & 27.14 & 26.85 & 27.32 & \textbf{27.50} \\
\hline\hline
     & 2 & 0.8687 & -& 0.8993 & 0.9026 & 0.9004 & 0.9056 & \textbf{0.9067}  \\
SSIM & 3 & 0.7736 & 0.7954 & 0.8076 & 0.8132 & 0.8093 & 0.8188 & \textbf{0.8215} \\
     & 4 & 0.7019 & -& 0.7331 & 0.7419 & 0.7352 & 0.7491 & \textbf{0.7513} \\
\hline\hline
     & 2 & 6.09 & -& 7.59 & 6.83 & 7.81 & \textbf{8.11} & 7.76 \\
IFC  & 3 & 3.41 & 2.98 & 4.14 & 3.83 & 4.23 & \textbf{4.45} & 4.26 \\
     & 4 & 2.23 & -& 2.71 & 2.57 & 2.78 & \textbf{2.94} & 2.74 \\
\hline\hline
     & 2 & 40.98 & -& 41.34 & 38.86 & 41.79 & \textbf{42.61} & 38.95 \\
NQM  & 3 & 33.15 & 29.06 & 37.12 & 35.23 & 37.22 & \textbf{38.24} & 35.25 \\
     & 4 & 26.15 & -& 31.17 & 29.18 & 31.27 & \textbf{32.31} & 30.46 \\
\hline\hline
     & 2 & 47.64 & -& 54.47 & 53.85 & 54.57 & \textbf{55.62} & 55.39 \\
WPSNR  & 3 & 39.72 & 41.66 & 43.22 & 43.56 & 43.36 & 44.25 & \textbf{44.32} \\
     & 4 & 35.71 & -& 37.75 & 38.26 & 37.85 & 38.72 & \textbf{38.87} \\
\hline\hline
     & 2 & 0.9813 & -& 0.9886 & 0.9890 & 0.9888 & 0.9896 & \textbf{0.9897} \\
MSSSIM  & 3 & 0.9512 & 0.9595 & 0.9643 & 0.9653 & 0.9647 & 0.9669 & \textbf{0.9675} \\
     & 4 & 0.9134 & -& 0.9317 & 0.9338 & 0.9326 & 0.9371 & \textbf{0.9376} \\
\hline
\end{tabular}
\end{center}
\vspace{-0.25cm}
\end{table*}

\begin{table*}[p]\footnotesize
\caption{The average results of PSNR (dB), SSIM, IFC, NQM, WPSNR (dB) and MSSIM on the BSD200 dataset.}\label{BSD200}
\vskip -0.35cm
\begin{center}
\begin{tabular}{|c|c|c|c|c|c|c|c|c|}
\hline
Eval. Mat & Scale& Bicubic & SC~\cite{Yang2010a} &NE+LLE~\cite{Chang2004} & KK~\cite{Kim2010} & ANR~\cite{Timofte2013} & A+~\cite{Timofte2013} & SRCNN\\
\hline\hline
     & 2 & 28.38 & - & 29.67 & 30.02 & 29.72 & 30.14 & \textbf{30.29} \\
PSNR & 3 & 25.94 & 26.54 & 26.67 & 26.89 & 26.72 & 27.05 & \textbf{27.18}  \\
     & 4 & 24.65 & -& 25.21 & 25.38 & 25.25 & 25.51 & \textbf{25.60} \\
\hline\hline
     & 2 & 0.8524 & -& 0.8886 & 0.8935 & 0.8900 & 0.8966 & \textbf{0.8977}  \\
SSIM & 3 & 0.7469 & 0.7729 & 0.7823 & 0.7881 & 0.7843 & 0.7945 & \textbf{0.7971} \\
     & 4 & 0.6727 & -& 0.7037 & 0.7093 & 0.7060 & 0.7171 & \textbf{0.7184}  \\
\hline\hline
     & 2 & 5.30 & -& 7.10 & 6.33 & 7.28 & \textbf{7.51} & 7.21 \\
IFC  & 3 & 3.05 & 2.77 & 3.82 & 3.52 & 3.91 & \textbf{4.07} & 3.91 \\
     & 4 & 1.95 & -& 2.45 & 2.24 & 2.51 & \textbf{2.62} & 2.45  \\
\hline\hline
     & 2 & 36.84 & -& 41.52 & 38.54 & 41.72 & \textbf{42.37} & 39.66 \\
NQM  & 3 & 28.45 & 28.22 & 34.65 & 33.45 & 34.81 & \textbf{35.58} & 34.72 \\
     & 4 & 21.72 & -& 25.15 & 24.87 & 25.27 & \textbf{26.01} & 25.65  \\
\hline\hline
     & 2 & 46.15 & -& 52.56 & 52.21 & 52.69 & 53.56 & \textbf{53.58} \\
WPSNR  & 3 & 38.60 & 40.48 & 41.39 & 41.62 & 41.53 & 42.19 & \textbf{42.29} \\
     & 4 & 34.86 & -& 36.52 & 36.80 & 36.64 & 37.18 & \textbf{37.24}  \\
\hline\hline
     & 2 & 0.9780 & -& 0.9869 & 0.9876 & 0.9872 & 0.9883 & \textbf{0.9883} \\
MSSSIM  & 3 & 0.9426 & 0.9533 & 0.9575 & 0.9588 & 0.9581 & 0.9609 & \textbf{0.9614} \\
     & 4 & 0.9005 & -& 0.9203 & 0.9215 & 0.9214 & 0.9256 & \textbf{0.9261} \\
\hline
\end{tabular}
\end{center}
\vspace{-0.25cm}
\end{table*}

\subsubsection{Running time}
\label{sec:running_time}

\begin{figure}[t]
\centering
  \includegraphics[width=\linewidth]{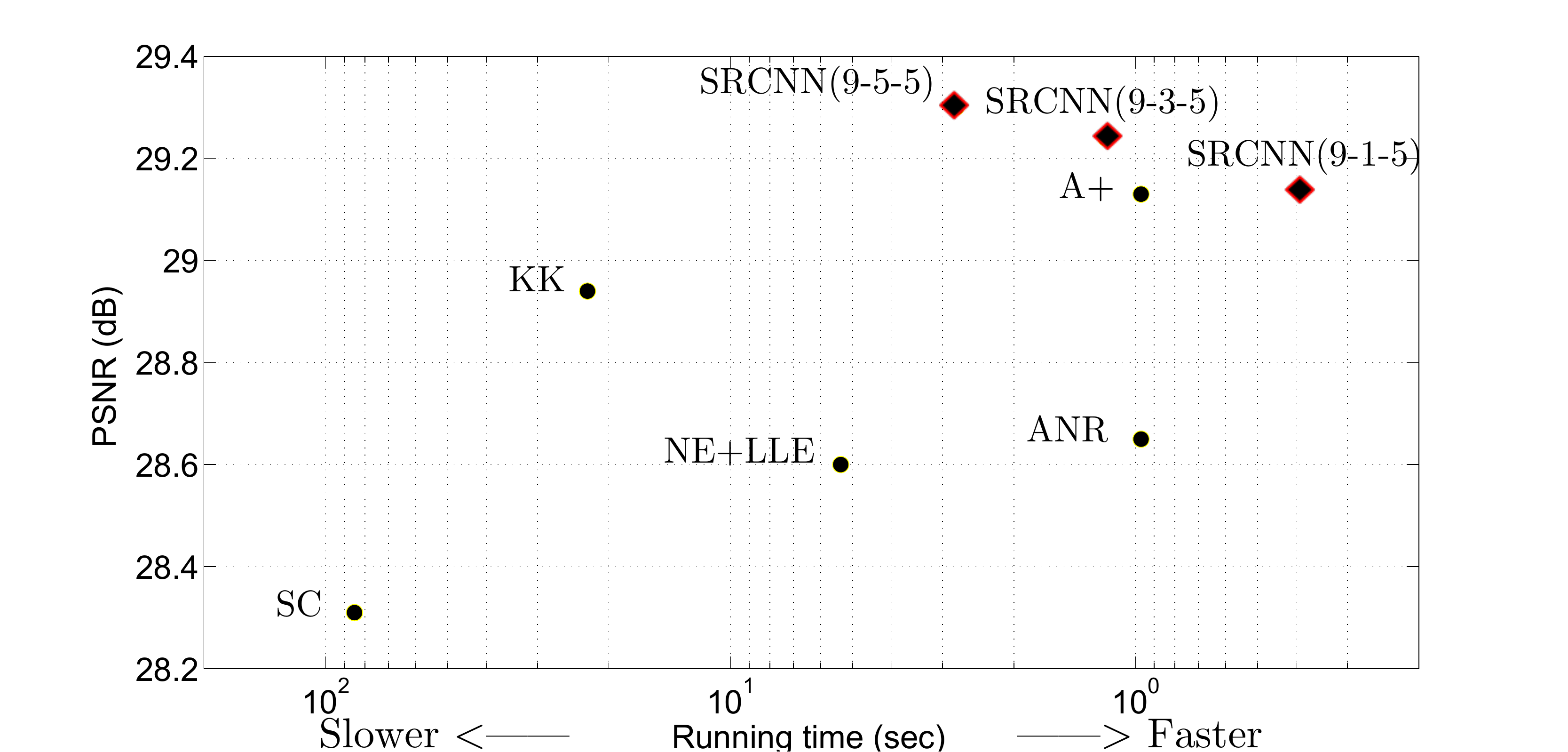}
\vskip -0.25cm
\caption{The proposed SRCNN achieves the state-of-the-art super-resolution quality, whilst maintains high and competitive speed in comparison to existing external example-based methods. The chart is based on Set14 results summarized in Table~\ref{set14}. The implementation of all three SRCNN networks are available on our project page.}
  \label{fig:runtime}
  \vskip -0.4cm
\end{figure}

Figure~\ref{fig:runtime} shows the running time comparisons of several state-of-the-art methods, along with their restoration performance on \textit{Set14}. All baseline methods are obtained from the corresponding authors' MATLAB+MEX implementation, whereas ours are in pure C++. We profile the running time of all the algorithms using the same machine (Intel CPU 3.10 GHz and 16 GB memory).
Note that the processing time of our approach is highly linear to the test image resolution, since all images go through the same number of convolutions.
Our method is always a trade-off between performance and speed. To show this, we train three networks for comparison, which are 9-1-5, 9-3-5, and 9-5-5. It is clear that the 9-1-5 network is the fastest, while it still achieves better performance than the next state-of-the-art A+.
Other methods are several times or even orders of magnitude slower in comparison to 9-1-5 network. Note the speed gap is not mainly caused by the different MATLAB/C++ implementations; rather, the other methods need to solve complex optimization problems on usage (e.g., sparse coding or embedding), whereas our method is completely feed-forward.
The 9-5-5 network achieves the best performance but at the cost of the running time.
The test-time speed of our CNN can be further accelerated in many ways, \eg~approximating or simplifying the trained networks~\cite{denton2014exploiting,jaderberg2014speeding,mamalet2012simplifying}, with possible slight degradation in performance.

\subsection{Experiments on Color Channels}
\label{sec:color}

In previous experiments, we follow the conventional approach to super-resolve color images. Specifically, we first transform the color images into the YCbCr space. The SR algorithms are only applied on the Y channel, while the Cb , Cr channels are upscaled by bicubic interpolation.
It is interesting to find out if super-resolution performance can be improved if we jointly consider all three channels in the process.

Our method is flexible to accept more channels without altering the learning mechanism and network design. In particular, it can readily deal with three channels simultaneously by setting the input channels to $c=3$.
In the following experiments, we explore different training strategies for color image super-resolution, and subsequently evaluate their performance on different channels.

\begin{table}[t]
\caption{Average PSNR (dB) of different channels and training strategies on the Set5 dataset.}\label{color_results}
\vspace{-0.25cm}
\begin{center}
\begin{tabular}{|c|c|c|c|c|}
\hline
Training &\multicolumn{4}{c|}{PSNR of different channel(s)}\\
 \cline{2-5}
Strategies & Y &  Cb &  Cr & RGB color image \\
\hline\hline
Bicubic & 30.39 & 45.44 & 45.42 & 34.57  \\
Y only & \textbf{32.39} & 45.44 & 45.42 & 36.37  \\
YCbCr & 29.25 & 43.30 & 43.49 & 33.47 \\
Y pre-train & 32.19 & \textbf{46.49} & \textbf{46.45} & 36.32 \\
CbCr pre-train & 32.14 & 46.38 & 45.84 & 36.25\\
RGB & 32.33 & 46.18 & 46.20 & \textbf{36.44} \\
KK & 32.37 & 44.35 & 44.22 & 36.32\\
\hline
\end{tabular}
\end{center}
\vskip -0.4cm
\end{table}

\noindent
\textbf{Implementation details.}
Training is performed on the 91-image dataset, and testing is conducted on the \textit{Set5}~\cite{Bevilacqua2012}. The network settings are: $c=3$, $f_1=9$, $f_2=1$, $f_3=5$, $n_1=64$, and $n_2=32$. As we have proved the effectiveness of SRCNN on different scales, here we only evaluate the performance of upscaling factor 3.

\noindent
\textbf{Comparisons.} We compare our method with the state-of-art color SR method -- KK~\cite{Kim2010}.
We also try different learning strategies for comparison:
\begin{packed_itemize}
\item \textbf{Y only:} this is our baseline method, which is a single-channel ($c=1$) network trained only on the luminance channel. The Cb, Cr channels are upscaled using bicubic interpolation.
\item \textbf{YCbCr:} training is performed on the three channels of the YCbCr space.
\item \textbf{Y pre-train:} first, to guarantee the performance on the Y channel, we only use the MSE of the Y channel as the loss to pre-train the network. Then we employ the MSE of all channels to fine-tune the parameters.
\item \textbf{CbCr pre-train:} we use the MSE of the Cb, Cr channels as the loss to pre-train the network, then fine-tune the parameters on all channels.
\item \textbf{RGB:} training is performed on the three channels of the RGB space.
\end{packed_itemize}

\begin{figure}[t]\tiny
\centering
\subfigure[First-layer filters -- Cb channel]{
  \label{deep1}
  \includegraphics[width=0.9\linewidth]{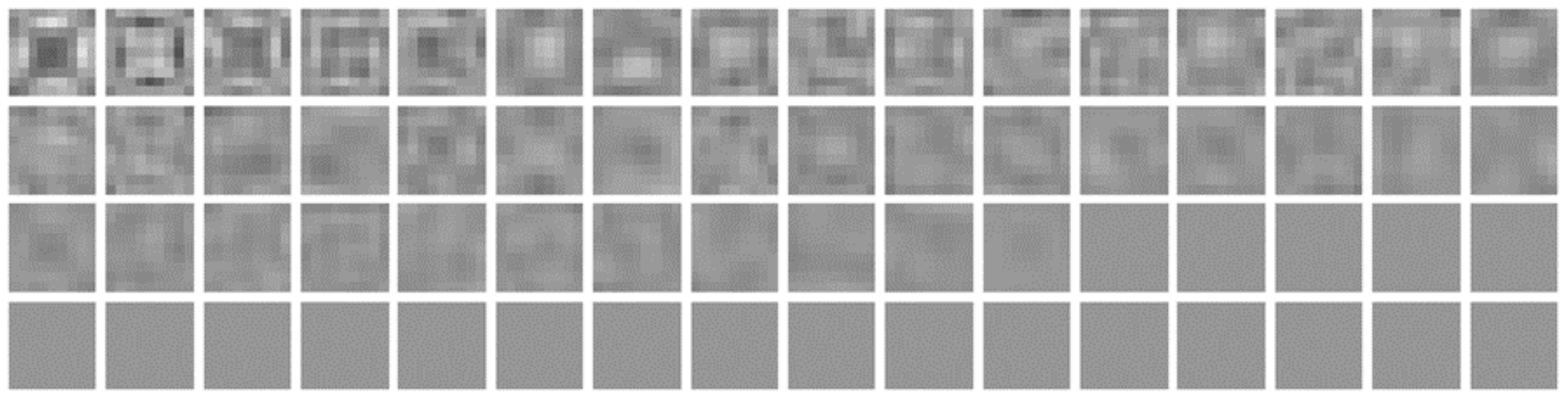}
}
\vskip -0.2cm
\subfigure[First-layer filters -- Cr channel]{
  \label{deep2}
  \includegraphics[width=0.9\linewidth]{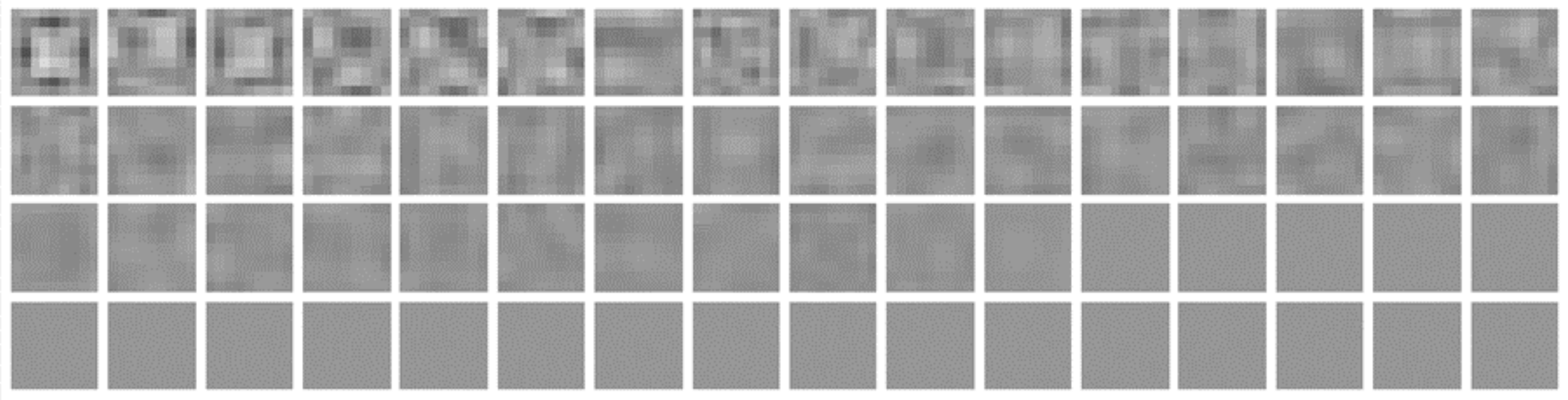}
}
\vskip -0.25cm
  \caption{Chrominance channels of the first-layer filters using the ``Y pre-train'' strategy.}\label{fig:color_channels}
\vskip -0.4cm
\end{figure}

The results are shown in Table~\ref{color_results}, where we have the following observations. (i) If we directly train on the YCbCr channels, the results are even worse than that of bicubic interpolation. The training falls into a bad local minimum, due to the inherently different characteristics of the Y and Cb, Cr channels.
(ii) If we pre-train on the Y or Cb, Cr channels, the performance finally improves, but is still not better than ``Y only'' on the color image (see the last column of Table~\ref{color_results}, where PSNR is computed in RGB color space). This suggests that the Cb, Cr channels could decrease the performance of the Y channel when training is performed in a unified network.
(iii) We observe that the Cb, Cr channels have higher PSNR values for ``Y pre-train'' than for ``CbCr pre-train''. The reason lies on the differences between the Cb, Cr channels and the Y channel. Visually, the Cb, Cr channels are more blurry than the Y channel, thus are less affected by the downsampling process. When we pre-train on the Cb, Cr channels, there are only a few filters being activated. Then the training will soon fall into a bad local minimum during fine-tuning. On the other hand, if we pre-train on the Y channel, more filters will be activated, and the performance on Cb, Cr channels will be pushed much higher. Figure~\ref{fig:color_channels} shows the Cb, Cr channels of the first-layer filters with ``Y pre-train'', of which the patterns largely differ from that shown in Figure~\ref{fig:patterns}.
(iv) Training on the RGB channels achieves the best result on the color image. Different from the YCbCr channels, the RGB channels exhibit high cross-correlation among each other. The proposed SRCNN is capable of leveraging such natural correspondences between the channels for reconstruction. Therefore, the model achieves comparable result on the Y channel as ``Y only'', and better results on Cb, Cr channels than bicubic interpolation.
(v) In KK~\cite{Kim2010}, super-resolution is applied on each RGB channel separately. When we transform its results to YCbCr space, the PSNR value of Y channel is similar as ``Y only'', but that of Cb, Cr channels are poorer than bicubic interpolation. The result suggests that the algorithm is biased to the Y channel.
On the whole, our method trained on RGB channels achieves better performance than KK and the single-channel network (``Y only''). It is also worth noting that the improvement compared with the single-channel network is not that significant (\ie~0.07 dB). This indicates that the Cb, Cr channels barely help in improving the performance.

\section{Conclusion}

We have presented a novel deep learning approach for single image super-resolution (SR). We show that conventional sparse-coding-based SR methods can be reformulated into a deep convolutional neural network. The proposed approach, SRCNN, learns an end-to-end mapping between low- and high-resolution images, with little extra pre/post-processing beyond the optimization.
With a lightweight structure, the SRCNN has achieved superior performance than the state-of-the-art methods. We conjecture that additional performance can be further gained by exploring more filters and different training strategies. Besides, the proposed structure, with its advantages of simplicity and robustness, could be applied to other low-level vision problems, such as image deblurring or simultaneous SR+denoising. One could also investigate a network to cope with different upscaling factors.

\bibliographystyle{splncs03}
\small{\bibliography{long,cnn_sr}}

\begin{IEEEbiography}[{\includegraphics[width=1in,height=1.25in,clip,keepaspectratio]{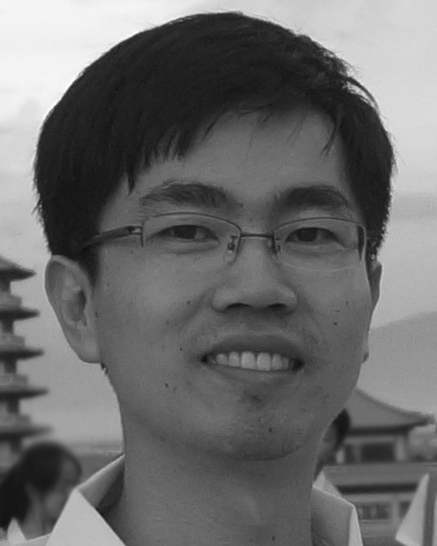}}]{Chao Dong}
received the BS degree in Information
Engineering from  Beijing Institute of Technology, China, in
2011. He is currently working toward the PhD
degree in the Department of Information Engineering
at the Chinese University of Hong Kong.
His research interests include image super-resolution and
denoising.
\end{IEEEbiography}
\vspace{-0.5cm}

\begin{IEEEbiography}[{\includegraphics[width=1in,height=1.25in,clip,keepaspectratio]{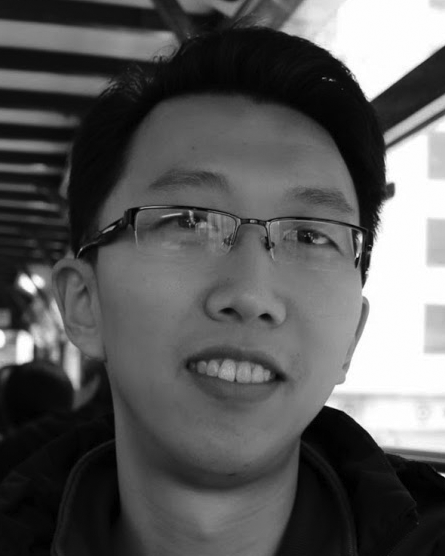}}]{Chen Change Loy}
received the PhD degree in Computer Science from the Queen Mary University of London in 2010. He is currently a Research Assistant Professor in the Department of Information Engineering, Chinese University of Hong Kong. Previously he was a postdoctoral researcher at Vision Semantics Ltd. His research interests include computer vision and pattern recognition, with focus on face analysis, deep learning, and visual surveillance.
\end{IEEEbiography}

\vspace{-0.5cm}

\begin{IEEEbiography}[{\includegraphics[width=1in,height=1.25in,clip,keepaspectratio]{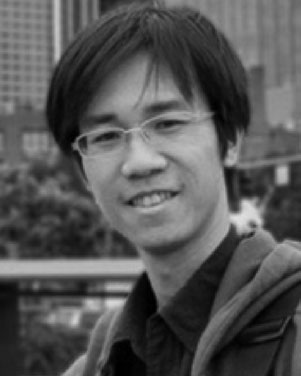}}]{Kaiming He}
 received the BS degree from
Tsinghua University in 2007, and the PhD degree
from the Chinese University of Hong Kong in
2011. He is a researcher at Microsoft Research
Asia (MSRA). He joined Microsoft Research Asia
in 2011. His research interests include computer
vision and computer graphics. He has won the
Best Paper Award at the IEEE Conference on
Computer Vision and Pattern Recognition
(CVPR) 2009. He is a member of the IEEE.
\end{IEEEbiography}

\vspace{-0.5cm}
\begin{IEEEbiography}[{\includegraphics[width=1in,height=1.25in,clip,keepaspectratio]{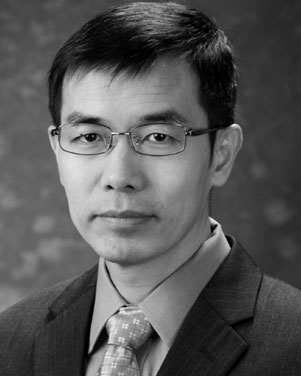}}]{Xiaoou Tang}
(S¡¯93-M¡¯96-SM¡¯02-F¡¯09) received
the BS degree from the University of Science
and Technology of China, Hefei, in 1990, the MS
degree from the University of Rochester, New
York, in 1991, and the PhD degree from the
Massachusetts Institute of Technology, Cambridge,
in 1996. He is a professor in the Department
of Information Engineering and an
associate dean (Research) of the Faculty of
Engineering of the Chinese University of Hong
Kong. He worked as the group manager of the
Visual Computing Group at the Microsoft Research Asia, from 2005 to
2008. His research interests include computer vision, pattern recognition,
and video processing. He received the Best Paper Award at the
IEEE Conference on Computer Vision and Pattern Recognition (CVPR)
2009. He was a program chair of the IEEE International Conference on
Computer Vision (ICCV) 2009 and he is an associate editor of the IEEE
Transactions on Pattern Analysis and Machine Intelligence and the
International Journal of Computer Vision. He is a fellow of the IEEE.
\end{IEEEbiography}

\vfill

\end{document}